\documentclass{article}

\usepackage{epsfig}
\usepackage{amsmath}
\usepackage{placeins}
\usepackage{url}
\usepackage{authblk}
\usepackage{orcidlink}
\usepackage{algorithm}
\usepackage{algpseudocode}
\usepackage{caption}
\usepackage{subcaption}
\usepackage{hyperref}
\usepackage{comment}
\usepackage{amssymb}
\usepackage{footnote}
\usepackage{natbib}
\usepackage{xfrac}

\newtheorem{definition}{Definition}
\newcommand{\email}[1]{\href{mailto:#1}{#1}}

\title{Elastic‑Net Multiple Kernel Learning: Combining Multiple Data Sources for Prediction}

\author[1,2,*]{Janaina Mourão-Miranda\orcidlink{0000-0002-3309-8441}}
\author[3]{Zakria Hussain}
\author[3]{Konstantinos Tsirlis}
\author[4]{Christophe Phillips\orcidlink{0000-0002-4990-425X}}
\author[1]{John Shawe-Taylor}
\affil[1]{Centre for Artificial Intelligence, Department of Computer Science, University College London, UK}
\affil[2]{Hawkes Institute, University College London, UK}
\affil[3]{Department of Computer Science, University College London, UK}
\affil[4]{GIGA - CRC Human Imaging, University of Liège, Belgium}
\affil[*]{\small Corresponding author: \email{j.mourao-miranda@ucl.ac.uk}}

\date{}

\begin{document}

\maketitle

\begin{abstract}

    Multiple Kernel Learning (MKL) models combine several kernels in supervised and unsupervised settings to integrate multiple data representations or sources, each represented by a different kernel. MKL seeks an optimal linear combination of base kernels that maximizes a generalized performance measure under a regularization constraint. Various norms have been used to regularize the kernel weights, including $l1$, $l2$ and $lp$, as well as the ``elastic-net" penalty, which combines $l1$- and $l2$-norm to promote both sparsity and the selection of correlated kernels. This property makes elastic-net regularized MKL (ENMKL) especially valuable when model interpretability is critical and kernels capture correlated information, such as in neuroimaging. Previous ENMKL methods have followed a two-stage procedure: fix kernel weights, train a support vector machine (SVM) with the weighted kernel, and then update the weights via gradient descent, cutting-plane methods, or surrogate functions. Here, we introduce an alternative ENMKL formulation that yields a simple analytical update for the kernel weights. We derive explicit algorithms for both SVM and kernel ridge regression (KRR) under this framework, and implement them in the open-source Pattern Recognition for Neuroimaging Toolbox (PRoNTo). We evaluate these ENMKL algorithms against $l1$-norm MKL and against SVM (or KRR) trained on the unweighted sum of kernels across three neuroimaging applications. Our results show that ENMKL matches or outperforms $l1$-norm MKL in all tasks and only underperforms standard SVM in one scenario. Crucially, ENMKL produces sparser, more interpretable models by selectively weighting correlated kernels.
    
\end{abstract}
\newpage


\section{Introduction}
\label{sec:intro}

Kernel methods were introduced to tackle learning tasks such as classification (e.g. support vector machines, SVM), regression (e.g. kernel ridge regression, KRR) and dimensionality reduction (e.g. kernel principal component analysis, KPCA) \cite{shawe2004kernel}. These methods represent data implicitly via a kernel function that encodes the similarity between data points and that can be linear or nonlinear. Multiple kernel learning (MKL) frameworks later emerged to combine several kernels, each capturing a different notion of similarity or derived from distinct information sources (\cite{Lanckriet2004, Bach2004, argyriou2005learning, Micchelli2005}). In essence, MKL algorithms search for the optimal linear combination of base kernels, subject to a chosen regularization, that maximizes a generalized performance criterion.

Multiple kernel learning (MKL) methods are particularly advantageous for integrating heterogeneous data sources that have very large numbers of features but relatively small sample sizes. By learning an optimal weighted linear combination of kernel functions, MKL allows each data source or feature group to contribute through a dedicated kernel while controlling overall model complexity through regularization of the kernel weights during training. This modular formulation reduces the risk of overfitting that commonly affects traditional machine learning approaches and many deep learning models in high-dimensional, low-sample-size settings. Empirical studies in multi-omics and other high-dimensional domains report that MKL approaches often yield better predictive performance than single kernel models or naive feature concatenation, particularly when sample sizes are limited \cite{briscik2024supervised}; for background on MKL algorithms and theory see \cite{gonen2011multiple}.

Several approaches have been proposed to regularize the kernel weights (or coefficients) in MKL, including $l1$- and  $l2$-norm regularized MKL (e.g., \cite{Rakotomamonjy2008, Kloft2008}), and a generalization to arbitrary $lp$-norm regularization with $ p \ge 1 $  \cite{Kloft2011}. Imposing sparsity via an $l1$-norm on the kernel combination is helpful for identifying the appropriate mix of data sources or feature subsets in many real-world applications. However, $l1$-norm MKL may discard correlated information, leading to suboptimal generalization performance and difficult model interpretation. Both $l2$- and $lp$-norm regularized MKL avoid this issue but yield non-sparse solutions and can incur high computational and storage costs.

To address the limitations described above, algorithms for elastic-net MKL (ENMKL), i.e. combining $l1$- and $l2$-norm regularisation, were proposed by Zheng-Peng and Xue-Gong (2011) \cite{Zheng-Peng2011}, Yang \emph{et al.} (2015) \cite{Yang2011} and Citi L. (2015) \cite{Citi2015} . In all cases the algorithm uses the standard two-stage technique of fixing the weights of the kernels, training an SVM with the weighted kernel and then updating the kernel weights. However, they differ on the strategy for updating the kernels weights. Zheng-Peng and Xue-Gong (2011) \cite{Zheng-Peng2011} used a gradient descent approach to update the kernel weights. The approach proposed by Yang et al. (2015) \cite{Yang2011} is based on the level method (which is a cutting plane method). The approach proposed in Citi L. (2015) \cite{Citi2015} is based on majorization-minimization, i.e. at each step it optimizes the kernel weights by minimizing a designed surrogate function, called majorizer. The main limitation of these algorithms is the complexity of the step to update the kernels weights. Furthermore, the ENMKL formulations in these papers are only derived for the SVM. We address these limitations by proposing an alternative elastic-net MKL formulation that leads to a simple analytical update for the kernel weights and present its derivations for the SVM and the KRR.

In this technical report we describe a two-stage ENMKL algorithm that is computationally efficient to solve. The algorithm can be applied to several kernel methods, and we show its use for the SVM and the KRR. We demonstrate the benefits of the proposed algorithm in three neuroimaging applications, including classification and regression problems, where multiple kernels correspond to different regions of the brain. The proposed ENMKL algorithm (with SVM and KRR) has been implemented in the Pattern Recognition for Neuroimaging Toolbox - PRoNTo v3.1 \url{http://www.mlnl.cs.ucl.ac.uk/pronto/} and can be applied to combine features from different brain regions as well as features from different data sources, such as imaging and non-imaging modalities for prediction. 

The rest of the report is organized as follows. In section \ref{sec:met} we introduce the methods. In section \ref{sec:experiments} we describe the experiments. In section \ref{sec:res} we present the results. Finally, in section \ref{sec:disc} we present a brief discussion and conclusion.


\section{Methods}
\label{sec:met}

\subsection{Preliminaries} 
\label{sec:prelim}

Let $\mathbf{z} = \{(x_i,y_i)\}_{i=1}^n$ be the learning set, where $x_i  \in \mathcal{X} \subset  \mathbb{R}^p$ and $y_i \in \mathcal {Y}$, with $ \mathcal {Z} = \mathcal{X} \times \mathcal {Y}$. For classification, $\mathcal{Y}=\{-1,1\}$; for regression, $\mathcal{Y}\subset\mathbb{R}$. 
Denote the input vectors by $\mathbf{x}=\{x_1,\dots,x_n\}$.

\begin{definition}[Aizerman \emph{et al.} (1964) \cite{Aizerman1964}]\label{def0}
A kernel is a function $\kappa$ that for all $x,x'\in\mathcal{X}$ satisfies
\[
  \kappa(x,x') \;=\;\bigl\langle \phi(x),\,\phi(x')\bigr\rangle,
\]
where 
\[
  \phi\colon \mathcal{X}\;\longrightarrow\;\mathcal{H}
\]
is a mapping into an inner‐product Hilbert space $\mathcal{H}$.
\end{definition}

In supervised learning the aim is to learn the relationship between $\mathbf{x}$ and $y$ by estimating a function $f: \mathbb{R}^p \rightarrow \ \mathcal {Y}$ such that, for every $x_i \in \mathcal{X} , f(x_i)$ provides the prediction of $y_i$ given $x_i$.  Kernel learning algorithms \cite{shawe2004kernel} make use of the $n \times n$  kernel matrix $\mathcal{K} = [\kappa (x_i,x_{i^{\prime}})]_{i,i^{\prime}=1}^n$ defined using the training inputs $\mathbf{x}$ and the solution of the learning problem can be written as
\begin{equation} 
\label{eq:kernel_alg}
\begin{split}
f(x)=\sum_{i=1}^n \alpha_i \kappa (x_i, x)+b
\end{split}
\end{equation}
where $\mathbf{\alpha}=(\alpha_1,\dots,\alpha_n)$ are dual weights and $b$ is a bias.
The pair $(\mathbf{\alpha},b)$ is found via methods such as SVM (Definition~\ref{def1})
or ridge regression (Definition~\ref{def7}).

For multiple sources or views of the data, the kernel (or kernel function) $\kappa (x,x')$ can be expressed as a convex combination of basis kernels:

\begin{equation} 
\label{eq:mkernel_alg}
\begin{split}
\kappa (x,x')= \sum_{j=1}^m\beta_j\kappa _j(x,x') \text{, with } \beta_j \geq 0,  \sum_{j=1}^m \beta_j=1
\end{split}
\end{equation}
Learning both the coefficients $\alpha_i$ and the weights $\beta_j$ in a single optimisation problem is known as the Multiple Kernel Learning (MKL) problem.  MKL algorithms are typically modifications of the SVM algorithm and can be expressed as finding a good combination of kernels within the SVM optimisation problem.

\subsection {Support Vector Machine and SVM-MKL}\

In this section we present the SVM optimization problem, a brief review of previous MKL formulations and the proposed Elastic-Net MKL formulation for SVM.

Let $\mathbf{z} = \{\phi(x_i),y_i\}_{i=1}^n$ be an $n$-sample of input–output pairs, where inputs $\phi(x) \in \mathit {R}^p$ are mapped using the feature mapping $\phi$, and $y \in  \mathcal {Y}$. Let $w \in \mathbb{R}^p$ be a $p$-dimensional weight vector and let $\xi \in \mathbb{R}^n$ be an $n$-dimensional vector of slack variables.

\begin{definition} \label{def1}
Primal Support Vector Machine
\begin{equation} 
\label{eq:eqn_svm}
\begin{split}
\min\limits_{w \in \mathbb{R}^p,\;\xi \in \mathbb{R}^n,\; b\in \mathbb{R}} \;\frac{1}{2} \Vert w \Vert{_2^2} + C \Vert \xi \Vert{_1} \\
\text{s.t. } y_i(\langle w,\phi(x_i) \rangle +b) \geq 1 - \xi{_i},\; i = 1,\dots,n, \\
\xi{_i} \geq 0,\; i = 1,\dots,n, \\
\end{split}
\end{equation}
where $b \in \mathbb{R}$ is the bias, and $C \in \mathbb{R}$ is the penalty parameter.
\end{definition}

The primal SVM problem can be converted into the dual form by using the technique of Lagrange multipliers. Several efficient optimization methods have been proposed to solve the SVM dual problem \cite{Buttou2007}.

\subsubsection {Previous MKL formulations}

We proceed by looking for a good combination of kernels within the SVM optimisation problem, by setting up the following block-norm Multiple Kernel Learning problem by Bach et al. 2004 \cite{Bach2004}.

\begin{definition}
\label{def2}
Primal Block-norm MKL-SVM
\begin{equation} 
\label{eq:eqn_pbn_mkl}
\begin{split}
\min\limits_{w_j \in \mathbb{R}^p,\; \xi \in \mathbb{R}^n,\; b \in \mathbb{R}} \; \frac{1}{2} {\left(\sum_{j=1}^m \Vert  w_j \Vert_2 \right)}^2 + C \Vert \xi \Vert_1 \\
\text{s.t. } y_i \left(\sum_{j=1}^m \langle w_j,\phi_j(x_i) \rangle + b \right) \geq 1 - \xi_i,\; i = 1,\dots, n, \\
\xi_i \geq 0,\; i = 1,\dots, n \\
\end{split}
\end{equation}
where $w_j$ is the weight vector of the $j^{th}$ feature space $\phi_j$.\end{definition}

The block-norm penalty ${\left(\sum_{j=1}^m \Vert  w_j \Vert_2 \right)}^2$ encourages sparsity at the kernel level, corresponding to an $l_1$-norm regularisation over the kernel weightings. We can follow this formulation in order to present the following elastic-net MKL formulation:

\begin{definition} 
\label{def3}
Primal Block-norm Elastic-Net MKL SVM
\begin{equation} 
\label{eq:eqn_en_mkl}
\begin{split}
\min\limits_{w_j \in \mathbb{R}^p, \; \xi \in \mathbb{R}^n, \; b \in \mathbb{R}} \; \frac{\mu}{2} {\left( \sum_{j=1}^m   \Vert w_j \Vert_2 \right)}^2 + \frac {1-\mu}{2} \left(\sum_{j=1}^m\Vert w_j \Vert_2^2 \right) + C \Vert \xi \Vert_1 \\
\text{s.t. } y_i \left( \sum_{j=1}^m \langle w_j,\phi_j(x_i) \rangle + b \right) \geq 1 - \xi_i,\; i = 1,\dots, n, \\
\xi_i \geq 0,\; i = 1,\dots, n \\
\end{split}
\end{equation}
where $\mu \in [0,1]$.
\end{definition}

When $\mu \rightarrow 1$ we have an $l1$-norm regularisation, and when $\mu \rightarrow 0$ we have an $l2$-norm regularisation. This optimization is computationally difficult to solve since the first term of the objective function is not smooth, therefore we present the following MKL formulation instead, in order to formulate the elastic-net MKL.

\begin{definition}
\label{def4}
Primal $l1$ norm regularised MKL SVM
\begin{equation} 
\label{eq:eqn_pnr_mkl}
\begin{split}
\min\limits_{w_j \in \mathbb{R}^p, \; \lambda_j \in \mathbb{R}^p, \; \xi_i \in \mathbb{R}^n, \; b \in \mathbb{R}} \; \frac{1}{2} {\sum_{j=1}^m \frac{1}{\lambda_j}  \Vert w_j \Vert_2^2} + C \Vert \xi \Vert_1 \\
\text{s.t. } y_i \left(\sum_{j=1}^m \langle w_j,\phi_j(x_i) \rangle + b \right) \geq 1 - \xi_i,\; i = 1,\dots, n, \\
\xi_i \geq 0,\; i = 1,\dots, n, \\
\sum_{j=1}^p \lambda_j = 1
\end{split}
\end{equation}
\end{definition}

Rakotomamonjy et al. (2008) \cite{Rakotomamonjy2008} have proven that the optimisations given in Definitions \ref{def2} and \ref{def4} are equivalent. 
We can follow Definition  \ref{def4} and propose the following elastic-net MKL.

\subsubsection {Proposed MKL formulation}

\begin{definition}
\label{def5}
Primal Elastic-Net MKL SVM
\begin{equation} 
\label{eq:eqn_enmkl}
\begin{split}
\min\limits_{w_j \in \mathbb{R}^p, \;  \lambda_j \in \mathbb{R}^p, \; \xi \in \mathbb{R}^n, \; b \in \mathbb{R}} \; \frac{1}{2} \sum_{j=1}^m \frac{\sqrt{\mu}}{\lambda_j}  \Vert w_j \Vert_2^2 + \frac{1}{2} \sum_{j=1}^m( {1-\mu})\Vert w_j \Vert_2^2 + C \Vert \xi \Vert_1 \\
\text{s.t. } y_i \,\left(\sum_{j=1}^m \langle w_j,\phi_j(x_i) \rangle + b\right) \geq 1 - \xi_i,\; i = 1,\dots, n, \\
\xi_i \geq 0,\; i = 1,\dots, n \\
\sum_{j=1}^m \sqrt{\mu}\,\lambda_j = 1
\end{split}
\end{equation}
\end{definition}

We show below that Definition \ref{def3} and Definition \ref{def5} are equivalent. 

The Lagrangian of the problem in Definition \ref{def5} is
\begin{equation} 
\label{eq:Lagrangian}
\begin{split}
 L = & \: \frac{1}{2} \sum_{j=1}^m \frac{\sqrt{\mu}}{\lambda_j}  \Vert w_j \Vert_2^2 + \frac{1}{2} \sum_{j=1}^m( {1-\mu})\Vert w_j \Vert_2^2 + C \Vert \xi \Vert_1  \\
 & -\sum_{i=1}^n \alpha_i \left[ y_i \left( \sum_{j=1}^m \langle w_j,\phi_j(x_i) \rangle + b \right) - 1 +\xi_i \right]  \\
 & - \sum_{i=1}^n \gamma_i \xi_i + \nu \left( \sum_{j=1}^m \sqrt{\mu}\,\lambda_j - 1 \right)
\end{split}
\end{equation}

Setting to zero the derivatives of the Lagrangian with respect to $w_j$ and $\lambda_j$, we get the following:

\begin{equation} 
\label{eq:deriv1}
\begin{split}
\frac{\partial L}{\partial w_j} =\left( \frac{\sqrt{\mu}}{\lambda_j} +1-\mu\right)  w_j - \sum_{i=1}^n \alpha_i y_i \phi_j(x_i) = 0\\
w_j = \frac{1}{\left(\frac{\sqrt{\mu}}{\lambda_j} +1-\mu\right)}  \sum_{i=1}^n \alpha_i y_i \phi_j(x_i)
\end{split}
\end{equation}

\begin{equation} 
\label{eq:deriv2}
\begin{split}
\frac{\partial L}{\partial \lambda_j} = -\frac{1}{2} \frac{\sqrt{\mu}}{\lambda_j^2}\Vert w_j \Vert_2^2 + \nu\sqrt{\mu} = 0\\
\lambda_j = \frac{1}{\sqrt{2\nu}} \Vert w_j \Vert_2
\end{split}
\end{equation}

The constraint $ \sum_{j=1}^m \sqrt{\mu}\,\lambda_j = 1$ implies that $\nu=\frac{\mu}{2}(\sum_{j=1}^m \Vert w_j \Vert_2)^2 $. Replacing $\nu$ in equation (\ref{eq:deriv2}) gives:

\begin{equation} 
\label{eq:lamb}
\begin{split}
\lambda_j = \frac{ \Vert w_j \Vert_2}{\sqrt{\mu} \sum_{j=1}^m \Vert w_j \Vert_2} 
\end{split}
\end{equation}

Finally, replacing $\lambda_j$ into Definition \ref{def5} leads to Definition \ref{def3}.

The optimisation problem of Definition \ref{def5} can be solved using an alternate optimisation algorithm. In the first step, the problem is optimised with respect to $w_j$, $b$ and $\xi$ with $\lambda_j$ fixed. Then in the second step, $\lambda_j$ is updated according to equation (\ref{eq:lamb}), with $w_j$, $b$ and $\xi$  fixed. 

We can rewrite the optimisation problem of Definition \ref{def5} by making the following substitution:
\begin{equation} 
\label{eq:eqn_wt}
\begin{split}
\tilde{w_j}= \frac{w_j} {\sqrt{\beta_j}}
\end{split}
\end{equation}
where $\beta_j = \frac {1}{\frac{\sqrt{\mu}}{\lambda_j} +(1-\mu)}$ , which gives us:

\begin{definition}
\label{def6}
Re-substituted Primal Elastic-Net MKL SVM
\begin{equation} 
\label{eq:eqn_renmkl}
\begin{split}
\min\limits_{\tilde{w_j} \in \mathbb{R}^p, \; \xi \in \mathbb{R}^n, \; b\in \mathbb{R}} \;  \frac{1}{2} \sum_{j=1}^m \Vert \tilde{w_j} \Vert{_2^2} + C \Vert \xi \Vert{_1} \\
\text{s.t. } y_i\, \left(\sum_{j=1}^m \langle \tilde{w_j},\tilde{\phi_j}(x_i) \rangle +b\right) \geq 1 - \xi{_i},\; i = 1,\dots,n, \\
\xi{_i} \geq 0,\; i = 1,\dots,n, \\
\text {where } \tilde{\phi_j}(x_i) = \sqrt{\beta_j}{\phi_j}(x_i) 
\end{split}
\end{equation}
where $b \in \mathbb{R}$ is the bias, and $C \in \mathbb{R}$ is the penalty parameter.
\end{definition}

This looks similar to a standard SVM optimisation problem, but with a re-weighted feature space and $m$ block components.  Since Definition \ref{def3}  is equivalent to Definition \ref{def5},  the optimisation of $w_j$ for a fixed $\lambda_j$ can be achieved by solving a standard SVM with the kernel matrix $\kappa = \sum_{j=1}^m \beta_j \kappa_j $, where the solution $\tilde{w}$ has a block decomposition $\tilde{w} = (\tilde{w_1},\dots,\tilde{w_m})$ and the solution of the original elastic-net is given by $w_j = \sqrt{\beta_j}\,\tilde{w_j}$. 

The steps to recover the primal weights $w_j$ for the original problem (Definition \ref{def5}) from the solution to the re-substituted problem (Definition \ref{def6}) are presented in the Appendix. Recovering the primal weights is especially useful when employing a linear kernel ($\kappa(x,x') = \langle x, x' \rangle$) as this enhances the model’s interpretability.

The algorithm to solve the optimisation problem of Definition \ref{def6} is given below (Algorithm \ref{alg:ENMKL-SVM}). In line 4 the dual SVM is solved using the Sequential Minimal Optimization (SMO) decomposition algorithm \cite{fan2005working}.  Note that step 10 is included to normalize the kernel weights $\beta_j$ such that $\sum_{j=1}^m \beta_j=1$ and step 9 rescales the dual variables $\alpha_i$  to compensate for the kernel weights normalization.

\begin{algorithm}
\caption{Elastic-Net MKL SVM (ENMKL-SVM) }
\label{alg:ENMKL-SVM}
\begin{algorithmic}[1]
\State \textbf{Input}: 
Kernel matrices $\mathcal{K}_1  = [\kappa_1(x_i,x_{i^{\prime}})]_{i,i^{\prime}=1}^n,\ldots, \mathcal{K}_m = [\kappa_m(x_i,x_{i^{\prime}})]_{i,i^{\prime}=1}^n$, $\mu \in [0,1].$
\State set each $\beta_j = 1/m$;
\While{kernel weights  $\beta_j$ have not converged}
\State $\alpha,b \gets$ train SVM with $\mathcal{K} = \sum_{j=1}^m \beta_j \kappa_j $ to find the dual weights vector
\State $\Vert {w_j} \Vert{_2} \gets \beta_j \sqrt{\sum_{i,{i^{\prime}}=1}^n{\alpha_i \alpha_{i^{\prime}} y_i y_{i^{\prime}} \kappa_j(x_i,x_{i^{\prime}})}}$
\State $\lambda_j \gets \frac{ \Vert w_j \Vert_2}{\sqrt{\mu} \sum_{j=1}^m \Vert w_j \Vert_2} $
\State $\beta_j \gets \frac {1}{\frac{\sqrt{\mu}}{\lambda_j} +(1-\mu)}$
\EndWhile
\State $\alpha_i \gets \alpha_i \sum_{j=1}^m \beta_j  $
\State $\beta_j \gets \frac {\beta_j}{\sum_{j=1}^m \beta_j}$ 
\State \textbf{Output}: final weights $\beta = (\beta_1,\dots,\beta_m)$, dual weights $\alpha$ and $b$
\end{algorithmic}
\end{algorithm}

\subsection {Ridge Regression and Kernel Ridge Regression MKL}\

Although standard MKL algorithms have generally been proposed under the optimization problem of an SVM, Cortes et al (2009) \cite{Cortes2009} proposed an $l2$-norm regularized MKL problem using ridge regression. In this section, we show that kernel ridge regression (KRR) using multiple kernels can be solved in our elastic-net framework. The ridge regression problem can be written as:

\begin{definition}
\label{def7}
Primal Ridge Regression (RR)
\begin{equation} 
\label{eq:eqn_rr}
\begin{split}
\min\limits_{w \in \mathbb{R}^p, \; \xi_i \in \mathbb{R}} \; \frac{1}{2}\Vert w \Vert{_2^2} + \frac{C}{n} \sum_{i=1}^n\xi_i^2   \\
\text{s.t. } \langle w,\phi(x_i) \rangle -y_i =  - \xi{_i},\; i = 1,\dots,n, \\
\end{split}
\end{equation}
\end{definition}

The ridge regression formulation can be extended to learn from multiple kernels. The following form follows the $l1$-norm regularized approach given for SVM in Definition \ref{def4}.

\begin{definition}
\label{def8}
Primal $l1$-norm regularised MKL RR
\begin{equation} 
\label{eq:eqn_l1_rr_mkl}
\begin{split}
\min\limits_{w_j \in \mathbb{R}^p, \; \xi_i \in \mathbb{R}, \; \lambda_j \in \mathbb{R}} \; \frac{1}{2}\sum_{j=1}^m\frac{1}{\lambda_j} \Vert w_j \Vert{_2^2} + \frac{C}{n} \sum_{i=1}^n \xi_i^2 \\
\text{s.t. } \sum_{j=1}^m\ \langle w_j,\phi_j(x_i) \rangle -y_i =  - \xi{_i},\; i = 1,\dots,n, \\
\end{split}
\end{equation}
\end{definition}

However, we would like have an analogous formulation to the elastic-net optimisation given in Definition \ref{def5}.

\begin{definition}
\label{def9}
Primal Elastic-Net MKL RR
\begin{equation} 
\label{eq:eqn_en_rr_mkl}
\begin{split}
\min\limits_{w_j \in \mathbb{R}^p, \; \xi_i \in \mathbb{R}, \; \lambda_j \in \mathbb{R}} \; \frac{1}{2}\sum_{j=1}^m \left( \frac{\sqrt{\mu}}{\lambda_j} + 1-\mu \right) \Vert w_j \Vert{_2^2} + \frac{C}{n} \sum_{i=1}^n \xi_i^2 \\
\text{s.t. } \sum_{j=1}^m\ \langle w_j,\phi_j(x_i) \rangle -y_i =  - \xi{_i}, \; i = 1,\dots,n, \\
\sum_{j=1}^m \sqrt{\mu}\,\lambda_j = 1
\end{split}
\end{equation}
\end{definition}

The Lagrangian of the problem in Definition \ref{def9} is:
\begin{equation} 
\label{eq:Lagrangian2}
\begin{split}
 L = & \: \frac{1}{2}\sum_{j=1}^m \left( \frac{\sqrt{\mu}}{\lambda_j} +1-\mu\right) \Vert w_j \Vert_2^2 + \frac{C}{n} \sum_{i=1}^n \xi_i^2  \\
 & - \sum_{i=1}^n \alpha_i\left( \sum_{j=1}^m\ \langle w_j,\phi_j(x_i) \rangle - y_i + \xi{_i}\right) + \nu \left( \sum_{j=1}^m \sqrt{\mu}\,\lambda_j - 1 \right)
\end{split}
\end{equation}

Setting to zero the derivatives of the Lagrangian with respect to $w_j$ and $\lambda_j$ we obtain the following.
\begin{equation} 
\label{eq:deriv3}
\begin{split}
\frac{\partial L}{\partial w_j} =  \left( \frac{\sqrt{\mu}}{\lambda_j} +1-\mu\right)  w_j - \sum_{i=1}^n \alpha_i \phi_j(x_i) = 0\\
w_j = \frac{1}{\left(\frac{\sqrt{\mu}}{\lambda_j} +1-\mu\right)}  \sum_{i=1}^n \alpha_i \phi_j(x_i)
\end{split}
\end{equation}
\begin{equation} 
\label{eq:deriv4}
\begin{split}
\frac{\partial L}{\partial \lambda_j} = -\frac{1}{2} \frac{\sqrt{\mu}}{\lambda_j^2}\Vert w_j \Vert_2^2 + \nu\sqrt{\mu} = 0\\
\lambda_j = \frac{1}{\sqrt{2\nu}} \Vert w_j \Vert_2
\end{split}
\end{equation}

Similarly as shown before, the constraint $ \sum_{j=1}^m \sqrt{\mu}\,\lambda_j = 1$ implies that $\nu=\frac{\mu}{2}(\sum_{j=1}^m \Vert w_j \Vert_2)^2 $ and replacing $\nu$ in equation (\ref{eq:deriv4}) gives:

\begin{equation} 
\label{eq:lamb2}
\begin{split}
\lambda_j = \frac{ \Vert w_j \Vert_2}{\sqrt{\mu} \sum_{j=1}^m \Vert w_j \Vert_2} 
\end{split}
\end{equation}

As in Definition \ref{def6} we can re-write the optimisation problem of Definition \ref{def9} by making the following substitution

\begin{equation} 
\label{eq:eqn_wt2}
\begin{split}
\tilde{w_j}= \frac{w_j} {\sqrt{\beta_j}}
\end{split}
\end{equation}
where $\beta_j = \frac {1}{\frac{\sqrt{\mu}}{\lambda_j} +(1-\mu)}$ , which gives us:

\begin{definition}
\label{def10}
Re-substituted Primal Elastic-Net MKL RR
\begin{equation} 
\label{eq:eqn_rr2}
\begin{split}
\min\limits_{w_j \in \mathbb{R}^p, \; \xi_i \in \mathbb{R}} \; \frac{1}{2}\sum_{j=1}^m \Vert \tilde{w_j} \Vert{_2^2} + \frac{C}{n} \sum_{i=1}^n \xi_i^2 \\
\text{s.t. } \sum_{j=1}^m\ \langle \tilde{w_j},\tilde{\phi_j}(x_i) \rangle -y_i =  - \xi{_i}, \; i = 1,\dots,n,\\
\text {where } \tilde{\phi_j}(x_i) = \sqrt{\beta_j}{\phi_j}(x_i) 
\end{split}
\end{equation}
\end{definition}

The algorithm to solve the optimisation problem of Definition \ref{def10} is almost identical to the SVM version described in Algorithm \ref{alg:ENMKL-SVM} and is described in the pseudocode of Algorithm \ref{alg:ENMKL-KRR}. As in the previous algorithm, step 10 is included to normalize the kernel weights $\beta_j$ such that $\sum_{j=1}^m \beta_j=1$ and step 9 rescales the dual variables $\alpha_i$ to compensate for the kernel weights normalization. As before, the steps to recover the primal weights of the original problem (Definition \ref{def9}) from the solution of the re-substituted problem (Definition \ref{def10}) are presented in the appendix.

\begin{algorithm}
\caption{Elastic-Net MKL KRR (ENMKL-KRR)}
\label{alg:ENMKL-KRR}
\begin{algorithmic}[1]
\State \textbf{Input}: 
Kernel matrices $\mathcal{K}_1  = [\kappa_1(x_i,x_{i^\prime})]_{i,i^\prime=1}^n,\dots,\mathcal{K}_m = [\kappa_m(x_i,x_{i^\prime})]_{i,i^\prime=1}^n$, $\mu \in [0,1].$
\State set each $\beta_j = 1/m$;
\While{kernel weights $\beta_j$ have not converged}
\State $\alpha  \gets$ train KRR with $\mathcal{K} = \sum_{j=1}^m \beta_j \kappa_j $ to find the dual weights vector
\State $\Vert {w_j} \Vert{_2} \gets \beta_j \sqrt{\sum_{i,{i^{\prime}}=1}^n{\alpha_i \alpha_{i^{\prime}} \kappa_j(x_i,x_{i^{\prime}})}}$
\State $\lambda_j \gets \frac{ \Vert w_j \Vert_2}{\sqrt{\mu} \sum_{j=1}^m \Vert w_j \Vert_2} $
\State $\beta_j \gets \frac {1}{\frac{\sqrt{\mu}}{\lambda_j} +(1-\mu)}$
\EndWhile
\State $\alpha_i \gets \alpha_i \sum_{j=1}^m \beta_j $
\State $\beta_j \gets \frac {\beta_j}{\sum_{j=1}^m \beta_j}$
\State \textbf{Output}: final weights $\beta = (\beta_1,\dots,\beta_m)$ and dual weights $\alpha$
\end{algorithmic}
\end{algorithm}

\section{Experiments}
\label{sec:experiments}

We applied the proposed algorithm to three neuroimaging data sets consisting of two classification tasks and one regression task. In the context of neuroimaging-based prediction, MKL can be applied to investigate the relative contributions of different brain regions and/or modalities for prediction. Here we applied the MKL framework to learn the optimal combination of individual brain regions for each predictive task. 

\subsection{Data description}
\label{sec:data_desc} 

\textbf{OASIS data set}: The first data was of a subset of the OASIS database \url{http://www.oasis-brains.org/} consisting of structural Magnetic Resonance Images (sMRI) from 50 demented patients and 50 healthy controls matched by age and gender, with ages varying from 62 to 92 years. The data were pre-processed using SPM8 \url{http://www.fil.ion.ucl.ac.uk/spm/software/}. The images were segmented into grey and white matter, normalized into MNI space, and spatially smoothed with a Gaussian kernel with a full width at half maximum (FWHM) of [8 8 8] mm following standard procedures in SPM8. For the OASIS data the task was classifying structural MRI of demented patients versus healthy controls. 
\\ \newline
\textbf{Haxby data set}: The second data consisted of a block design functional Magnetic Resonance Images (fMRI) experiment acquired using a visual paradigm, where the participants passively viewed grey scale images of eight categories: pictures of faces, cats, houses, chairs, scissors, shoes, bottles, and control, non-sense images. We used data from a single subject (participant 1), consisting of 12 runs, each comprising eight blocks of 24s showing one of the eight different object types separated by periods of rest. Each image was shown for 500 ms followed by a 1500 ms inter-stimulus interval. Full-brain fMRI data were recorded with a volume repetition time of 2.5s. Each category block therefore corresponds approximately to nine scans, separated by six scans of rest. For further information on the acquisition parameters, please consult the original reference \cite{haxby2001distributed}. The data were pre-processed using SPM8. The images were motion corrected, segmented and normalized into MNI space. No smoothing was applied to the data. For the Haxby data the task was classifying fMRI volumes during the visualization of ‘faces’ versus visualization of ‘houses’.
\\ \newline
\textbf{IXI data set}: The third data consisted of a subset of the IXI database \url{https://brain-development.org/ixi-dataset/} consisting of sMRI volumes. We focused on the images of the Aged group (age range: 60 to 90) from Guys Hospital which has 102 subjects. The data were pre-processed in SPM8. The images were segmented into grey and white  matter, normalized into MNI space, and smoothed with a Gaussian kernel with a full width at half maximum (FWHM) of [10 10 10] mm following standard procedures in SPM8. For the IXI dataset the task was to predict the subject's age from their structural MR images.

\subsection{Experimental Protocol} 
\label{sec:exp_prot}

For each data set we used a predefined anatomical template (Automated Anatomical Labeling, AAL template \cite{tzourio2002automated})  to partition the images into 116 anatomically defined brain regions. The complete AAL atlas (including all regions) was used as a mask to select the voxels or features from each dataset. For each region we computed a linear kernel based on the regional pattern of brain tissue or activation contained in all voxels within the region. The kernels were mean-centered and normalized. Mean centering the kernel corresponds to mean centering the features across samples (i.e. it is equivalent to subtracting the mean of each feature/voxel, computing the mean based on the training data), while normalizing the kernel corresponds to dividing each feature vector (i.e. each sample) by its norm. The latter operation is important when using MKL approaches to compensate for the fact that the different kernels might be computed from different numbers of features (e.g. brain regions or modalities with different sizes).

The experiments were implemented using the PRoNTo software batch framework \cite{schrouff2013pronto}. We built new wrapper functions for the elastic-net MKL SVM algorithm (ENMKL-SVM) and for the elastic-net KRR algorithm (ENMKL-KRR). We compared both elastic-net algorithms (ENMKL-SVM and ENMKL-KRR) with the SimpleMKL, which corresponds to a $l1$-norm regularized MKL algorithm \cite{Rakotomamonjy2008}, and is implemented in PRoNTo (i.e. simpleMKL classification and simpleMKL regression). In addition we also compared the MKL classification models with an SVM trained on the unweighted sum of the kernels and the MKL regression models with a KRR trained on the unweighted sum of the kernels. We performed a nested k-fold cross-validation procedure (k=5 for the internal and external loop) for the OASIS and IXI datasets and a nested k-fold on the blocks for the Haxby dataset (k=6 for the internal and external loop). The external loop was used for assessing the models' performance (accuracy or mean squared error for classification and regression, respectively) and the internal loop was used for optimizing the models' hyperparameters. In table \ref{tab:hyper2} the range of hyperparameters explored by each model is presented.

\begin{table}[h!]
\caption{Hyperparameters range for $C$ in all models (SVM, KRR, SimpleMKL classification/regression, ENMKL-SVM/KRR) and $\mu$ in the ENMKL-SVM/KRR.}
\label{tab:hyper2}
\centering
\vspace{0.3cm}
\begin{tabular}{lcc}
Hyperparameter \hspace{0.5cm} & Range \hspace{0.5cm} \\
\hline
C & $10^{[-3 \; -2 \; -1 \; 0 \; 1 \; 2 \; 3]}$ \\
$\mu$  & [0.1 0.2 0.3 0.4 0.5 0.6 0.7 0.8 0.9 1]	\\
\hline
\end{tabular}
\end{table}

\section{Results} 
\label{sec:res}

We applied the protocol described in Section \ref{sec:exp_prot} to the data sets described in Section \ref{sec:data_desc}. 

\subsection {OASIS Results} 
\label{sec:OASIS}

Table \ref{tab:multi1} shows the performance and number of kernels (brain regions) used in the different classification models in the OASIS  dataset. The balanced accuracy and Area Under the Curve (AUC) were higher for the SVM than for both MKL models. Furthermore, the balanced accuracy was slightly higher for the ENMKL-SVM than the SimpleMKL. The ENMKL-SVM selected all kernels (116) while the SimpleMKL only selected 67 kernels. These results are in line with the fact that $l1$-norm regularisation provides a much sparser solution than the elastic-net regularisation, in particular when there are many correlated features/kernels. The results suggest that the discriminative patterns in the OASIS data was distributed over the whole brain as the models based on all kernels presented higher performance.

\begin{table}[h!]
\caption{OASIS dataset results}
\label{tab:multi1}
\centering
\vspace{0.3cm}
\begin{tabular}{lccc}
 & Balanced Accuracy \hspace{0.1cm} & AUC \hspace{0.1cm} & Number of kernels \hspace{0.1cm}\\
\hline
ENMKL-SVM & 71 & 75  & 116\\
SimpleMKL classification  & 66 & 75 & 67\\
\textbf{SVM} & \textbf{75} & \textbf{80} & \textbf{116} \\
\hline
AUC - Area Under the Curve
\end{tabular}
\end{table}

\begin{figure}[H]
    \centering
    \begin{subfigure}[b]{0.48\textwidth}
    \centering
    \includegraphics[width=\textwidth]{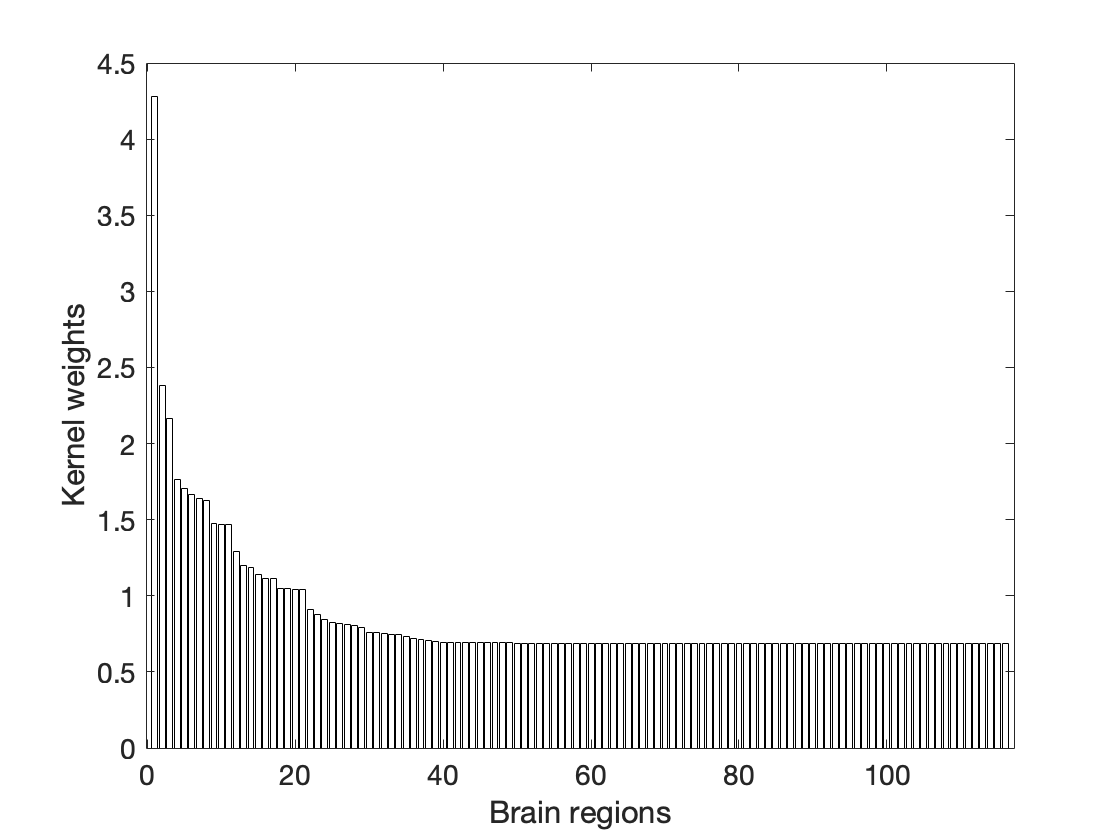}
    \caption{ENMKL-SVM}
    \label{fig:ENMKL_OASIS}
    \end{subfigure}
    ~
    \begin{subfigure}[b]{0.48\textwidth}
    \centering
    \includegraphics[width=\textwidth]{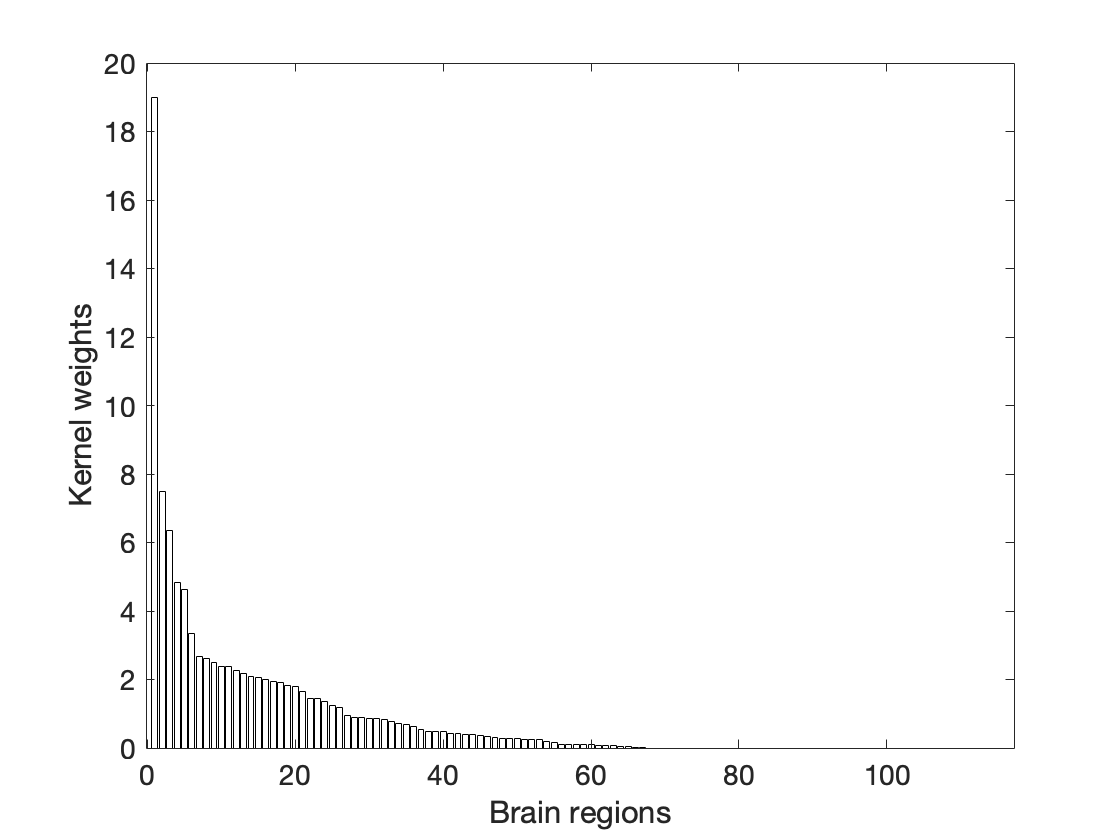}
    \caption{SimpleMKL}
    \label{fig:SimpleMKL_OASIS}
    \end{subfigure}
    
    \caption{Kernel weights distribution - OASIS data}
    \label{fig:image1}

\end{figure}

Figure  \ref{fig:image1} shows the distribution of kernel weights per brain regions for the ENMKL-SVM and SimpleMKL, and tables \ref{tab:list_regions1} and \ref{tab:list_regions2} show the top regions ranked according to the kernel weights for each model, respectively. The SimpleMKL provides a sparser solution and attributes higher weights for the top  kernels than the ENMKL-SVM.

\begin{table}[h!]
\caption{ENMKL-SVM: top 15 regions ranked according to the kernel weights for classifying demented patients versus healthy controls in the OASIS dataset.}
\label{tab:list_regions1}
\centering
\vspace{0.3cm}
\begin{tabular}{lcc}
Brain regions \hspace{0.3cm} & Kernel weight \hspace{0.3cm}  & Number of voxels \hspace{0.3cm} \\
\hline
Hippocampus Right	&4.28	&951\\
Postcentral Right	&2.38	&3846\\
Cerebellum 10 Left	&2.17	&148\\
Occipital Middle Right	&1.77	&1997\\
Insula Left	&1.71	&1898\\
Precentral Right	&1.66	&3375\\
Temporal Middle Left	&1.64	&4887\\
Cingulum Anterior Right	&1.63	&1286\\
SupraMarginal Left	&1.47	&1206\\
Frontal Middle Left	&1.47	&4802\\
Frontal Superior Left	&1.47	&3654\\
Parietal Superior Right	&1.29	&2143\\
Amygdala Right	&1.2	&240\\
Angular Left	&1.18	&1147\\
Cerebellum Crus 1 Right	&1.14	&2583\\
\hline
\end{tabular}
\end{table}

\begin{table}[h!]
\centering
\caption{SimpleMKL classification: top 15 regions ranked according to the  kernel weights for classifying demented versus healthy controls in the OASIS data set.}
\label{tab:list_regions2}
\vspace{0.3cm}
\begin{tabular}{lcc}
Brain regions \hspace{0.3cm} & Kernel weight \hspace{0.3cm}  & Number of voxels \hspace{0.3cm} \\
\hline
Amygdala Left	&18.99	 &211	\\
Frontal Middle Left	&7.5	&4802	\\
Hippocampus Right	&6.34	&951	\\
SupraMarginal Left	&4.84	&1206	\\
Thalamus Left	&4.62	&1056	\\
Temporal Middle Left	&3.35	&4887	\\
Insula Left	&2.66	&1898	\\
Lingual Left	&2.62	&2164	\\
Postcentral Right	&2.5	&3846	\\
Cerebellum Crus 1 Left	&2.39	&2550	\\
Rolandic Operculum Right 	&2.39	&1366	\\
Precuneus Right	&2.27	&3256	\\
Occipital Middle Right	&2.17	&1997	\\
Temporal Pole Superior Right	&2.08	&1315	\\
Occipital Superior Left	&2.06	&1317	\\
\hline
\end{tabular}
\end{table}

\clearpage

\subsection {Haxby Results} 
\label{sec:Haxby}

Table \ref{tab:multi2} shows the performance and number of kernels (brain regions) used in the different classification models in the Haxby dataset. The ENMKL-SVM and SimpleMKL models had the same performance (in terms of balanced accuracy and AUC) which was higher than the performance of the SVM model. Similarly to the previous example, the number of selected kernels by the ENMKL-SVM was higher than the number selected by the SimpleMKL, however fewer kernels were selected by both models in comparison with the previous task. These results indicate that the discriminative patterns in the Haxby dataset are more sparse than the discriminative pattern in the OASIS dataset, which makes sense with the considered classification problems, as dementia can affect the whole brain while visual processing engages a specific brain network with a smaller number of regions.

\begin{table}[h!]
\centering
\caption{Haxby dataset results.}
\label{tab:multi2}
\vspace{0.3cm}
\begin{tabular}{lccc}
 & Balanced Accuracy  \hspace{0.1cm} & AUC \hspace{0.1cm}  & Number of kernels \hspace{0.1cm}\\
\hline
\textbf{ENMKL-SVM} & \textbf{97.22} & \textbf{100} & \textbf{56}\\
\textbf{SimpleMKL classification} & \textbf{97.22}  & \textbf{100} & \textbf{18}\\
SVM  & 81.02 & 98 & 116 \\
\hline
AUC - Area Under the Curve
\end{tabular}
\end{table}

\begin{figure}[H]
\centering
\begin{subfigure}[b]{0.48\textwidth}
\centering
\includegraphics[width=\textwidth]{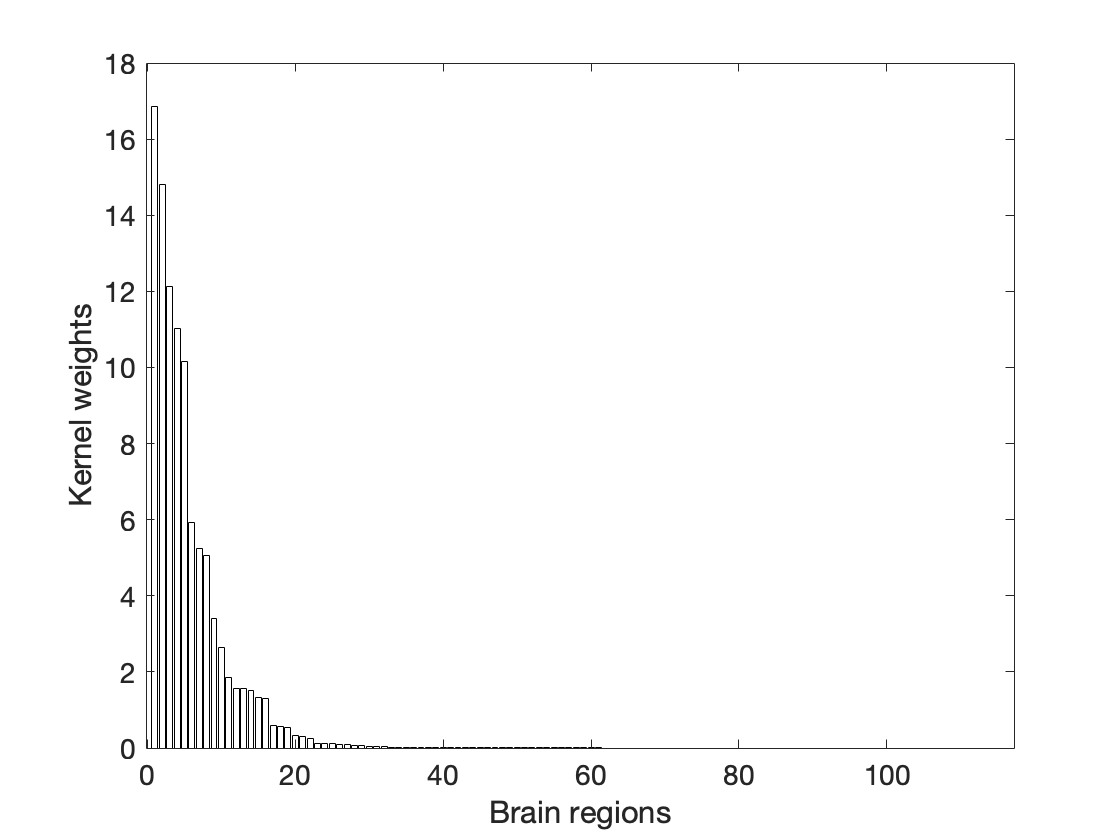}
\caption{ENMKL-SVM }
\label{fig:ENMKL_OASIS}
\end{subfigure}
~
\begin{subfigure}[b]{0.48\textwidth}
\centering
\includegraphics[width=\textwidth]{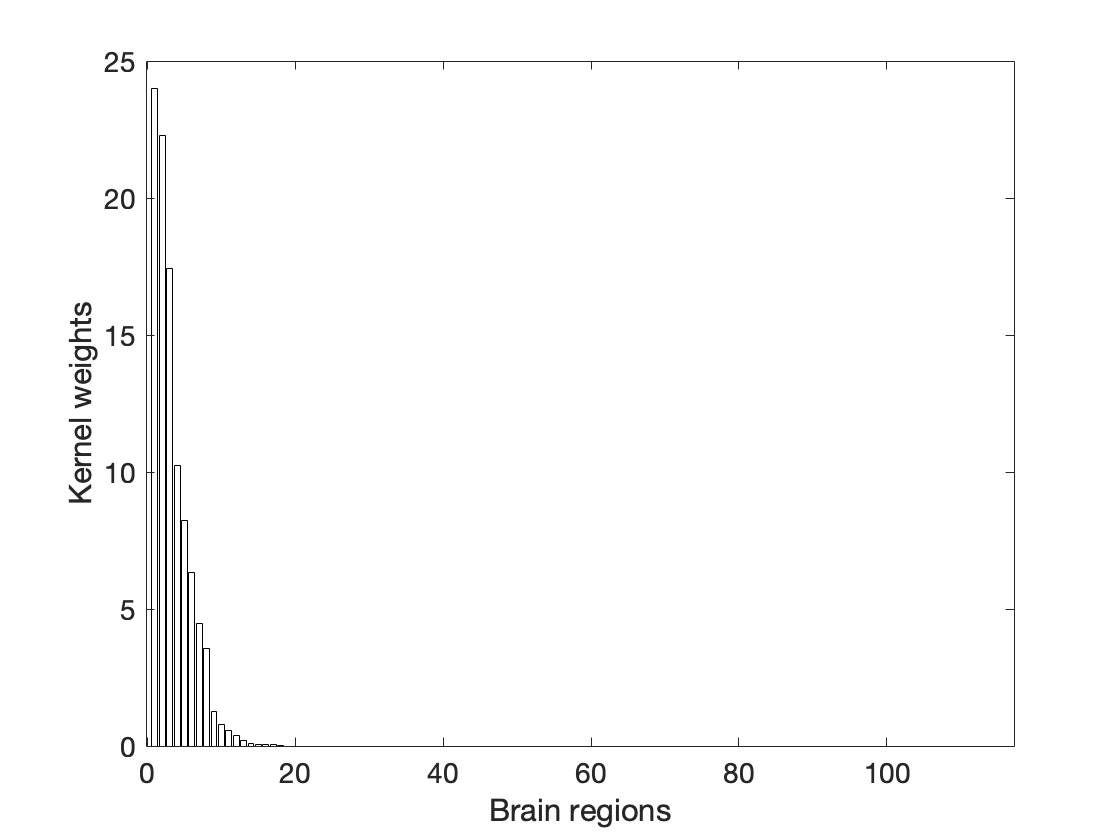}
\caption{SimpleMKL}
\label{fig:SimpleMKL_OASIS}
\end{subfigure}

\caption{Kernel weights distribution - Haxby data}
\label{fig:image2}

\end{figure}

Figure  \ref{fig:image2} shows the distribution of kernel weights per brain regions for the ENMKL-SVM and SimpleMKL, and tables \ref{tab:list_regions3} and \ref{tab:list_regions4} show the 15 top brain regions ranked according to the kernel weights for both models, respectively. The ENMKL-KRR model gives more similar weights for selected bilateral regions than the SimpleMKL model, which are likely to have correlated information.

\begin{table}[h!]
\centering
\caption{ENMKL-SVM: top 15 regions ranked according to the kernel weights for classifying fMRI images during the visualization of faces versus houses in the Haxby dataset.}
\label{tab:list_regions3}
\vspace{0.3cm}
\begin{tabular}{lcc}
Brain regions \hspace{0.3cm} & Kernel weight \hspace{0.3cm}  & Number of voxels \hspace{0.3cm} \\
\hline
\textbf{Fusiform Left}	&16.87	&664	\\	
\textbf{Lingual Right}	&14.81	&694	\\	
\textbf{Occipital Middle Left}	&12.12	&936	\\	
\textbf{Fusiform Right}	&11.02	&761	\\	
\textbf{Lingual Left} 	&10.16	&645	\\	
Cingulum Posterior Left	&5.93	&143	\\	
Parietal Superior Left	&5.25	&600	\\	
Frontal Inferior Operculum Right	&5.05	&407	\\	
Precuneus Left	&3.41	&1046	\\	
Caudate Left	&2.65	&285	\\
\textbf{Calcarine Left}	&1.85	&662	\\	
Occipital Superior Left	&1.57	&394	\\	
Hippocampus Right	&1.56	&284	\\	
\textbf{Calcarine Right}	&1.52	&	27	\\	
\textbf{Occipital Middle Right}	&1.32	&608	\\	
\hline
Bilateral brain regions in \textbf{bold}
\end{tabular}
\end{table}

\begin{table}[h!]
\centering
\caption{SimpleMKL classification: top 15 regions ranked according to the  kernel weights for classifying fMRI images during the visualization of faces versus houses in the Haxby dataset.}
\label{tab:list_regions4}
\vspace{0.3cm}
\begin{tabular}{lcc}
Brain regions \hspace{0.3cm} & Kernel weight \hspace{0.3cm}  & Number of voxels \hspace{0.3cm} \\
\hline
\textbf{Lingual Right}	&24	&694	\\
\textbf{Fusiform Left}	&22.29	 &64	\\
\textbf{Occipital Middle Left}	&17.42	 &936	\\
\textbf{Lingual Left}	&10.25	 &645	\\
\textbf{Fusiform Right}	&8.22	&761	\\
Cingulum Posterior Left &6.34&	143	\\
Frontal Inferior Operculum Right &4.47&	407	\\
Parietal Superior Left	&3.54	&600	\\
Caudate Left	&1.24	&285	\\
Hippocampus Right &0.79	&284	\\
Precuneus Left	&0.56	&1046	\\
\textbf{Occipital Middle Right}	&0.38	&608	\\
Insula Left	&0.22	&537	\\
Cingulum Anterior Left	&0.08	&402	\\
Cuneus Right	&0.07	&426	\\
\hline
Bilateral brain regions in \textbf{bold}
\end{tabular}
\end{table}

\subsection {IXI dataset} 
\label{sec:IXI}

Table \ref{tab:multi3} shows the performance and number of kernels used in the different regression models for predicting age in the IXI dataset. The lowest mean squared error (MSE) and higher correlation between the real and predicted age were obtained by the ENMKL-KRR, followed by KRR and SimpleMKL regression. All kernels were selected by the ENMKL-KRR while the SimpleMKL regression model selected 59 out of the 116 kernels. These results indicate that the predictive pattern is distributed, i.e. all brain regions seem to have predictive information about age. 

\begin{table}[h!]
\centering
\caption{IXI dataset results.}
\label{tab:multi3}
\vspace{0.3cm}
\begin{tabular}{lccc}
 & Mean Squared Error \hspace{0.1cm} & Correlation  \hspace{0.1cm} & Number of kernels \hspace{0.1cm}\\
\hline
\textbf{ENMKL-KRR} & \textbf{25.25} & \textbf{0.54} & \textbf{116}\\
SimpleMKL regression & 29.63 & 0.44  & 59\\
KRR & 29.18 &0.48 &116 \\
\hline
\end{tabular}
\end{table}

\begin{figure}[H]
\centering
\begin{subfigure}[b]{0.48\textwidth}
\centering
\includegraphics[width=\textwidth]{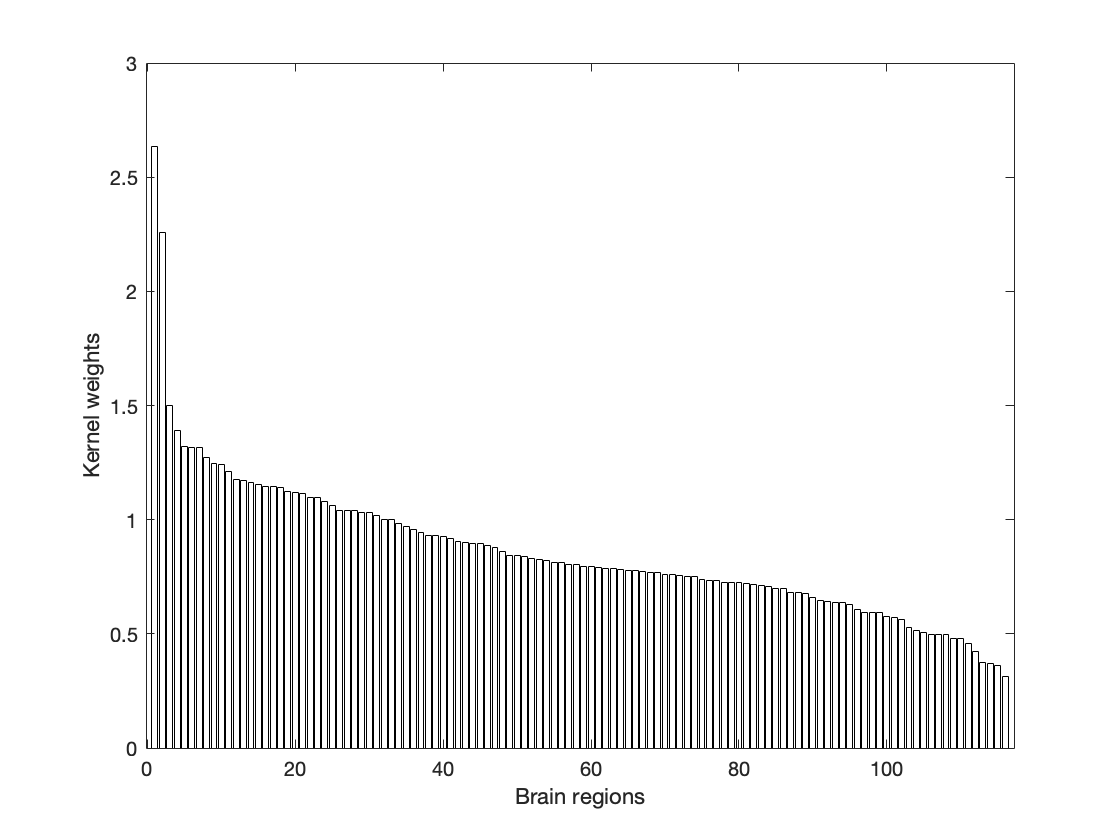}
\caption{ENMKL-SVM}
\label{fig:ENMKL_OASIS}
\end{subfigure}
~
\begin{subfigure}[b]{0.48\textwidth}
\centering
\includegraphics[width=\textwidth]{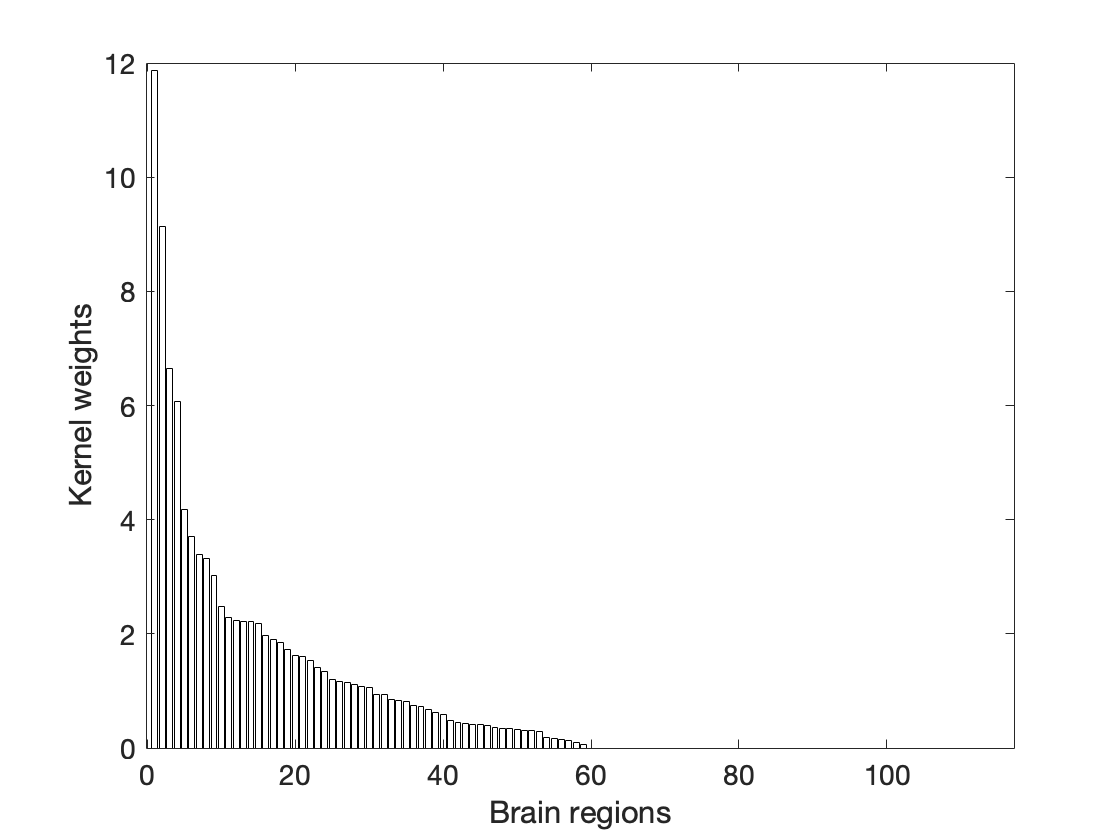}
\caption{SimpleMKL}
\label{fig:SimpleMKL_OASIS}
\end{subfigure}

\caption{Kernel weights distribution - IXI data}
\label{fig:image3}

\end{figure}

Figure  \ref{fig:image3} shows the distribution of kernel weights per brain regions for the ENMKL-SVM and SimpleMKL, and tables \ref{tab:list_regions5} and \ref{tab:list_regions6} show the top regions ranked according to the kernel weights for both models, respectively. The ENMKL-KRR selects more bilateral regions (within the top regions) than the SimpleMKL model.

\begin{table}[h!]
\centering
\caption{ENMKL-KRR: top 15 regions ranked according to the kernel weights for predicting age in the IXI dataset.}
\label{tab:list_regions5}
\vspace{0.3cm}
\begin{tabular}{lcc}
Brain regions \hspace{0.3cm} & Kernel weight \hspace{0.3cm}  & Number of voxels \hspace{0.3cm} \\
\hline
\textbf{Heschl L}	&2.63	&515 \\
\textbf{Heschl R}	&2.26	&536 \\
Hippocampus L	&1.50	&2208\\
Temporal Sup L	&1.39	&5360\\
\textbf{Fusiform L}	&1.32	&5253\\
Pallidum L	&1.32	&637\\
Putamen L	&1.31	&2240\\
Cingulum Post R	&1.27	&784\\
Caudate L	&1.24	&2227\\
Frontal Mid Orb R	&1.24	&2311\\
Frontal Inf Orb R 	&1.21	&4103\\
Frontal Mid Orb R	&1.17	&2020\\
Precentral R	&1.17	&7996\\
\textbf{Fusiform R}	&1.16	&5958\\
Occipital Sup R	&1.15	&3277\\
\hline
Bilateral brain regions in \textbf{bold}
\end{tabular}
\end{table}

\begin{table}[h!]
\centering
\caption{SimpleMKL regression: top 15 regions ranked according to the kernel weights for predicting age in the IXI dataset.}
\label{tab:list_regions6}
\vspace{0.3cm}
\begin{tabular}{lcc}
Brain regions \hspace{0.3cm} & Kernel weight \hspace{0.3cm}  & Number of voxels \hspace{0.3cm} \\
\hline
Heschl Left	&11.87	 &515	\\
\textbf{Frontal Middle Orbital Right}	&9.13	&2311	\\
Putamen Left	&6.65	&2240	\\
Hippocampus Left	&6.07	&2208	\\
Fusiform Left	&4.18	&5253	\\
Cerebellum 8 Left	&3.71	&2847	\\
\textbf{Frontal Middle Orbital Right}	&3.38	&2020	\\
Frontal Superior Right 	&3.31	&9336	\\
Occipital Superior Right	&3.03	&3277	\\
Occipital Inferior Left	&2.48	&2292	\\
Cingulum Anterior Right	&2.29	&3060	\\
Parietal Inferior Right	&2.23	&3155	\\
Cerebellum 6 Left	&2.22	&4131	\\
Caudate Left	&2.21	&2227	\\
Precuneus Left	&2.17	&8389	\\
\hline
Bilateral brain regions in \textbf{bold}
\end{tabular}
\end{table}

\clearpage

\FloatBarrier
\section{Discussion and Conclusion} 
\label{sec:disc}

In this report, we have described an elastic-net multiple kernel learning (ENMKL) algorithm that can be applied to several kernel learning methods and showed its derivations for the support vector machine and the ridge regression (i.e. ENMKL-SVM and ENMKL-KRR) models. We compared the performance of the proposed ENMKL algorithm with the SimpleMKL algorithm and baseline classification and regression models (SVM and KRR) in three different neuroimaging datasets. The applications included two classification tasks (classifying patterns of brain structure in demented patients vs healthy controls and classifying patterns of brain activation during visualization of faces versus houses) and a regression task (predicting age from patterns of brain structure). We showed that the ENMKL models presented similar or better performance than the SimpleMKL models in all tasks, although it presented lower accuracy than the SVM for classifying demented patients vs healthy controls. Furthermore, as expected, due to the combination of the $l_1$-norm and $l_2$-norm  regularisation, the ENMKL models were able to identify more brain regions with correlated information (e.g. bilateral regions) and provide insights on the level of sparsity of the predictive task.

We conclude the report by stating that, while we demonstrate the benefits of the proposed algorithm for neuroimaging applications using linear kernels, it can be applied to other multimodal or multi-source predictive problems that combine both  linear and non-linear kernels. The ENMKL algorithm is especially beneficial when one or more modalities (or sources) have far more features than examples, since its complexity scales with the number of examples rather than the number of features (it operates on the kernel matrix instead of the raw data matrix). 

Further work should compare the performance of the proposed ENMKL algorithm with other MKL and multi-source approaches.

\section*{Code and data availability}

All models reported here were implemented in the open‑access software PRoNTo v3.1 (http://www.mlnl.cs.ucl.ac.uk/pronto/). The PRoNTo release and instructions for reproducing our results will be made publicly available upon acceptance of this Technical Report. The datasets used in our experiments are accessible from the PRoNTo website.

\section*{Appendix}

In case of the linear kernel $\kappa(x,x') = \langle x, x' \rangle$, the feature mapping simplifies to $\phi(x)= x$. In this case, it is possible to recover the primal weights of the kernel learning algorithms from its dual weights, which enhance the models' interpretability.

Here we describe the steps to recover the primal weights $w_j$ of the original elastic-net MKL SVM problem (Definition \ref{def5}) from the solution to the re-substituted problem (Definition \ref{def6}). 

The Lagrangian of the alternative problem in Definition \ref{def6} is
\begin{equation} 
\label{eq:Lagrangian3}
\begin{split}
 L = & \: \frac{1}{2} \Vert \tilde{w_j} \Vert_2^2 + C \Vert \xi \Vert_1  \\
 & -\sum_{i=1}^n \tilde{\alpha_i} \left[ y_i \left( \sum_{j=1}^m \langle \tilde{w_j}, \tilde{\phi_j}(x_i) \rangle + b \right) - 1 +\xi_i \right]  \\
 & - \sum_{i=1}^n \gamma_i \xi_i 
\end{split}
\end{equation} 

Setting the derivatives of Lagrangian w.r.t. $\tilde{w_j}$ to zero gives:

\begin{equation} 
\label{eq:deriv5}
\begin{split}
\frac{\partial L}{\partial \tilde{w_j}} =  \tilde{w_j} - \sum_{i=1}^n \tilde{\alpha_i}  y_i \tilde{\phi_j}(x_i) = 0\\
\tilde{w_j} = \sum_{i=1}^n \tilde{\alpha_i} y_i \tilde{\phi_j}(x_i)
\end{split}
\end{equation}

Replacing $\tilde{\phi_j}(x_i) = \sqrt{\beta_j}\phi_j(x_i)$ into $\tilde{w_j}$ and $\tilde{w_j}$ into $w_j =\sqrt{\beta_j} \tilde{w_j}$ leads to $w_j= \beta_j \sum_{i=1}^n \tilde{\alpha_i} y_i\phi_j(x_i)$ where $\tilde{\alpha_i}$ and $\beta_j$ are outputs of algorithm \ref{alg:ENMKL-SVM}. In the linear case $w_j= \beta_j \sum_{i=1}^n \tilde{\alpha_i} y_i x_i$.

Similarly, we can recover the weights $w_j$ of the original elastic-net MKL RR problem (Definition \ref{def9}) from the solution to the re-substituted problem (Definition \ref{def10}). 

The Lagrangian of the alternative problem in Definition \ref{def10} is:

\begin{equation} 
\label{eq:Lagrangian4}
\begin{split}
 L = & \: \frac{1}{2}\Vert \tilde{w_j} \Vert_2^2 + \frac{C}{n} \sum_{i=1}^n \xi_i^2  \\
 & - \sum_{i=1}^n \alpha_i\left( \sum_{j=1}^m\ \langle \tilde{w_j},\tilde{\phi_j}(x_i) \rangle - y_i + \xi{_i}\right)
\end{split}
\end{equation}

Setting the derivatives of Lagrangian w.r.t. $\tilde{w_j}$ to zero gives:

\begin{equation} 
\label{eq:deriv6}
\begin{split}
\frac{\partial L}{\partial \tilde{w_j}} = \tilde{w_j} - \sum_{i=1}^n \alpha_i \tilde{\phi_j}(x_i) = 0\\
\tilde{w_j} =  \sum_{i=1}^n \alpha_i \tilde{\phi_j}(x_i)
\end{split}
\end{equation}

Replacing $\tilde{\phi_j}(x_i) = \sqrt{\beta_j}\phi_j(x_i)$ into $\tilde{w_j}$ and $\tilde{w_j}$ into $w_j =\sqrt{\beta_j} \tilde{w_j}$ leads to $w_j= \beta_j \sum_{i=1}^n \tilde{\alpha_i} \phi_j(x_i)$ where $\tilde{\alpha_i}$ and $\beta_j$ are outputs of algorithm \ref{alg:ENMKL-KRR}. In the linear case $w_j= \beta_j \sum_{i=1}^n \tilde{\alpha_i} x_i$.

\begin{figure}[H]
\centering
\begin{subfigure}[b]{0.48\textwidth}
\centering
\includegraphics[width=\textwidth]{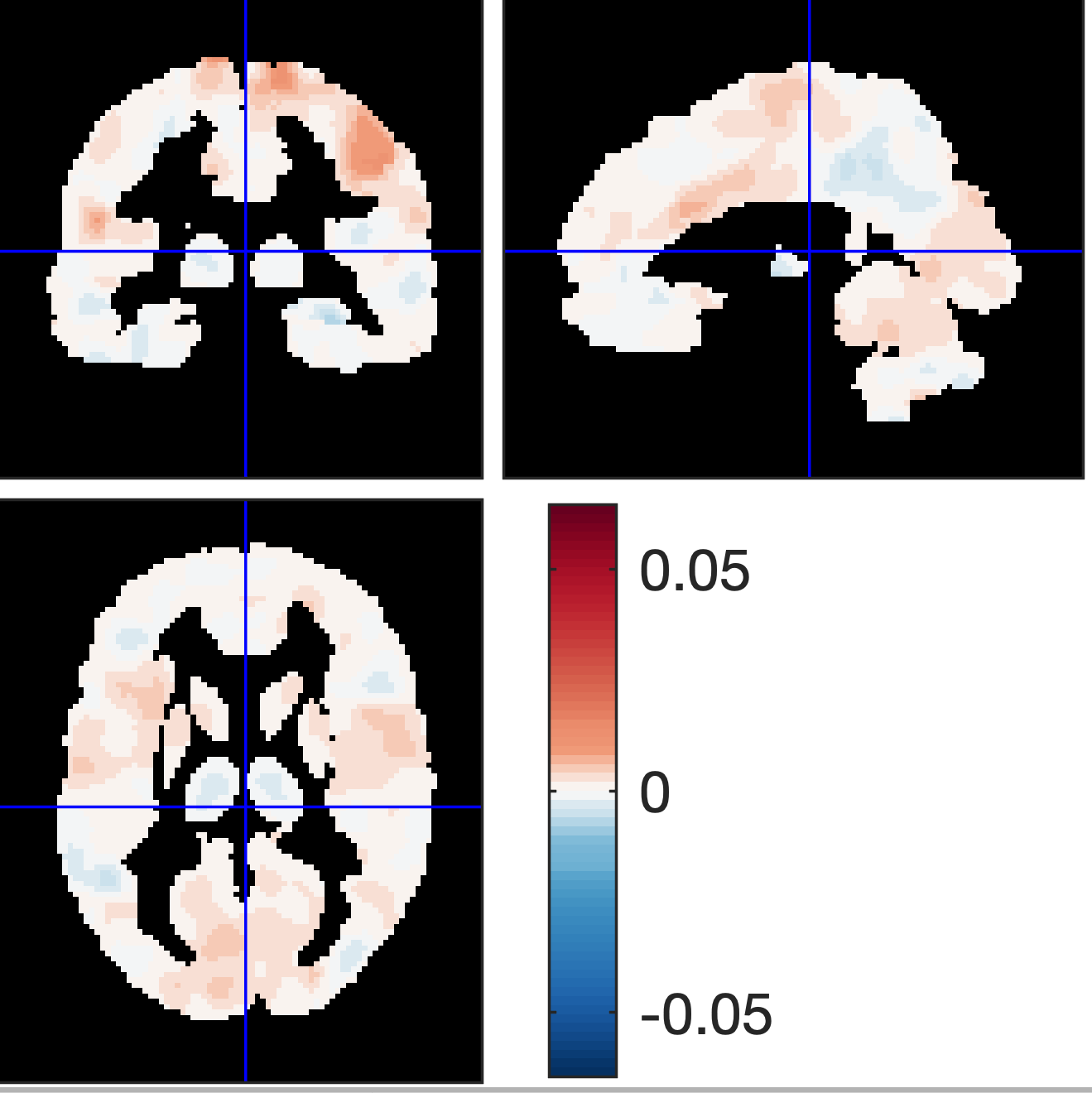}
\caption{ENMKL-SVM}
\end{subfigure}
~
\begin{subfigure}[b]{0.48\textwidth}
\centering
\includegraphics[width=\textwidth]{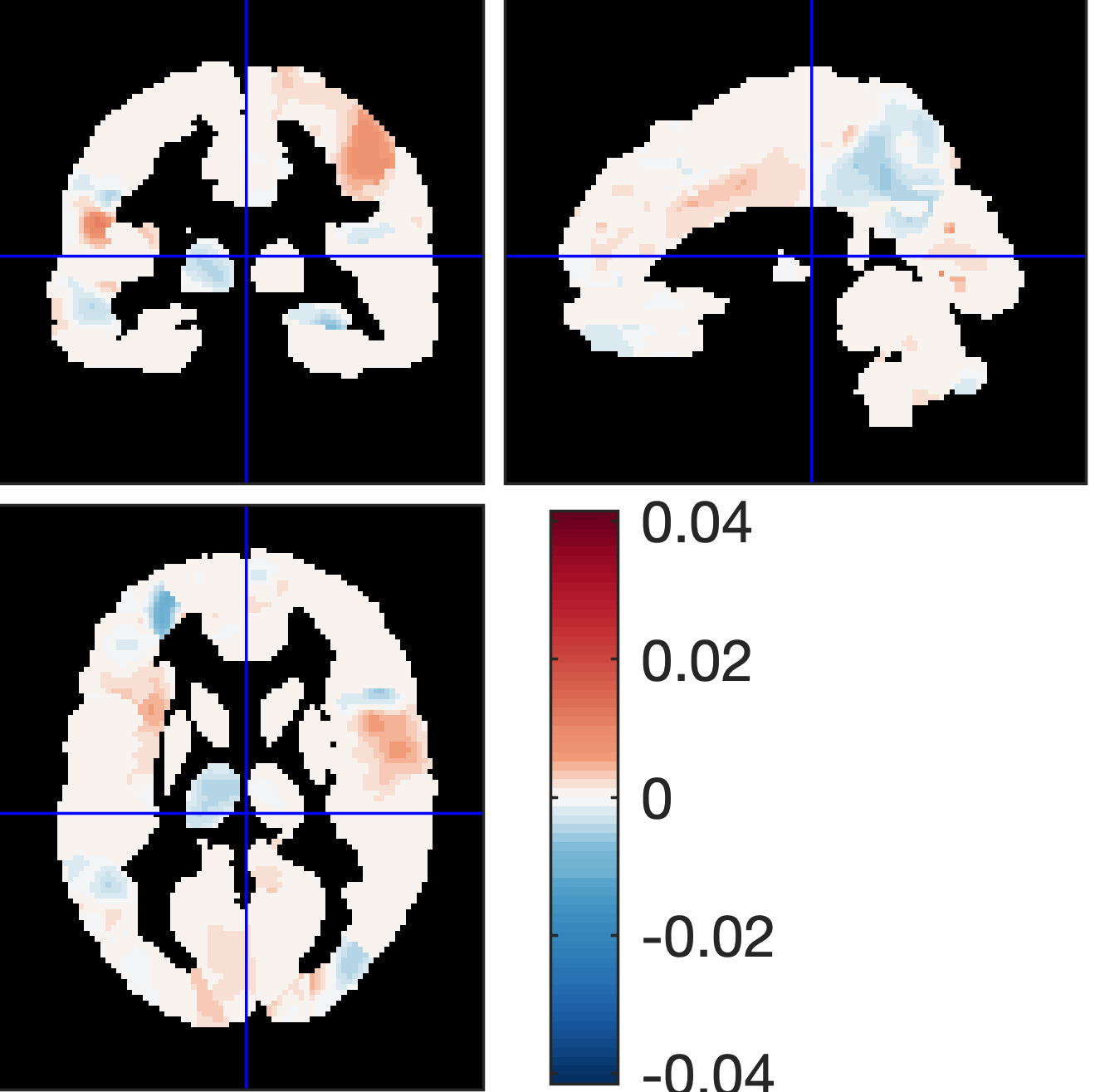}
\caption{SimpleMKL}
\end{subfigure}
~
\begin{subfigure}[b]{0.48\textwidth}
\centering
\includegraphics[width=\textwidth]{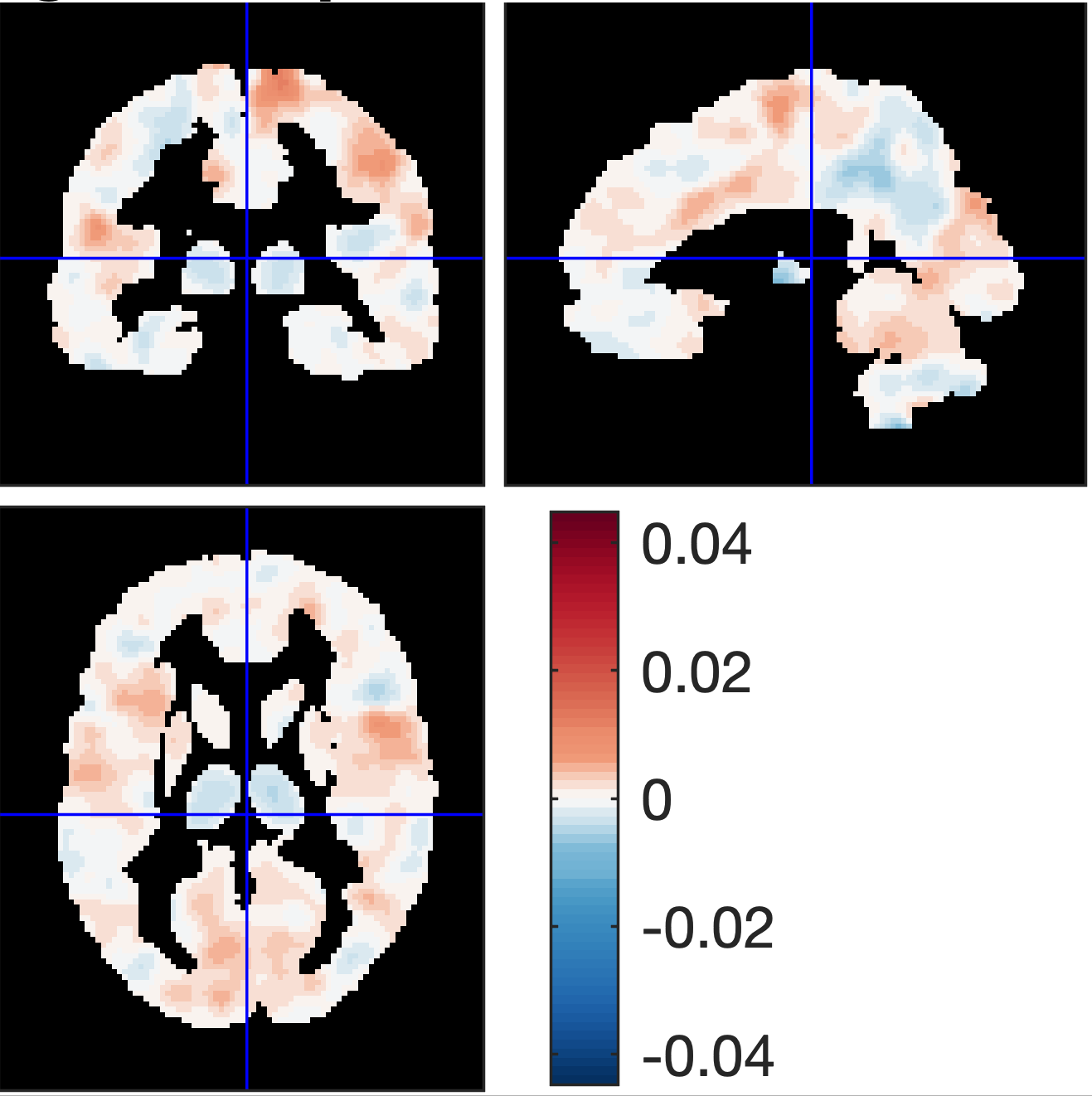}
\caption{SVM}
\end{subfigure}

\caption{Weight maps - OASIS data}
\label{fig:wmOASIS}

\end{figure}

\begin{figure}[H]
\centering
\begin{subfigure}[b]{0.48\textwidth}
\centering
\includegraphics[width=\textwidth]{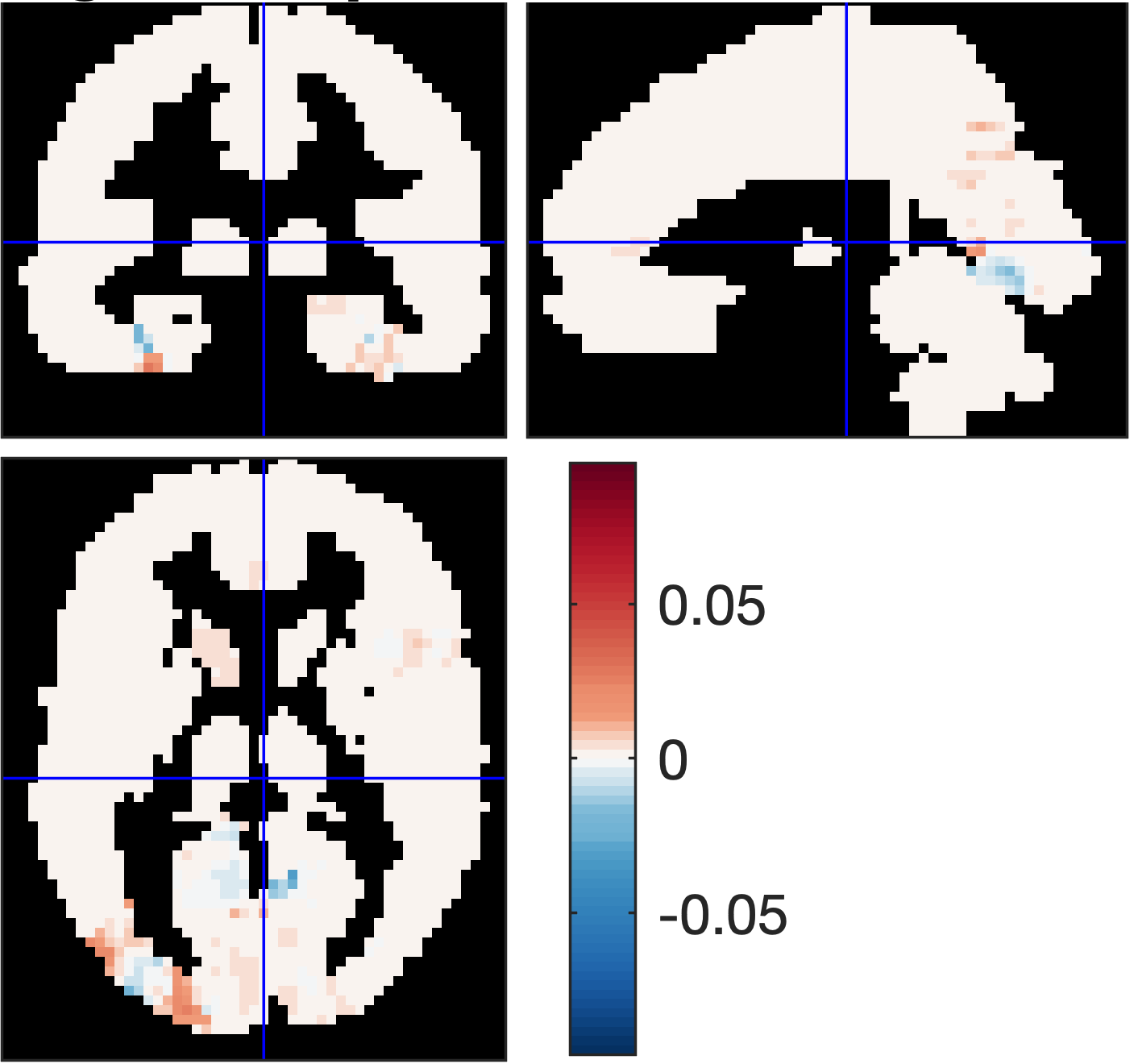}
\caption{ENMKL-SVM}
\end{subfigure}
~
\begin{subfigure}[b]{0.48\textwidth}
\centering
\includegraphics[width=\textwidth]{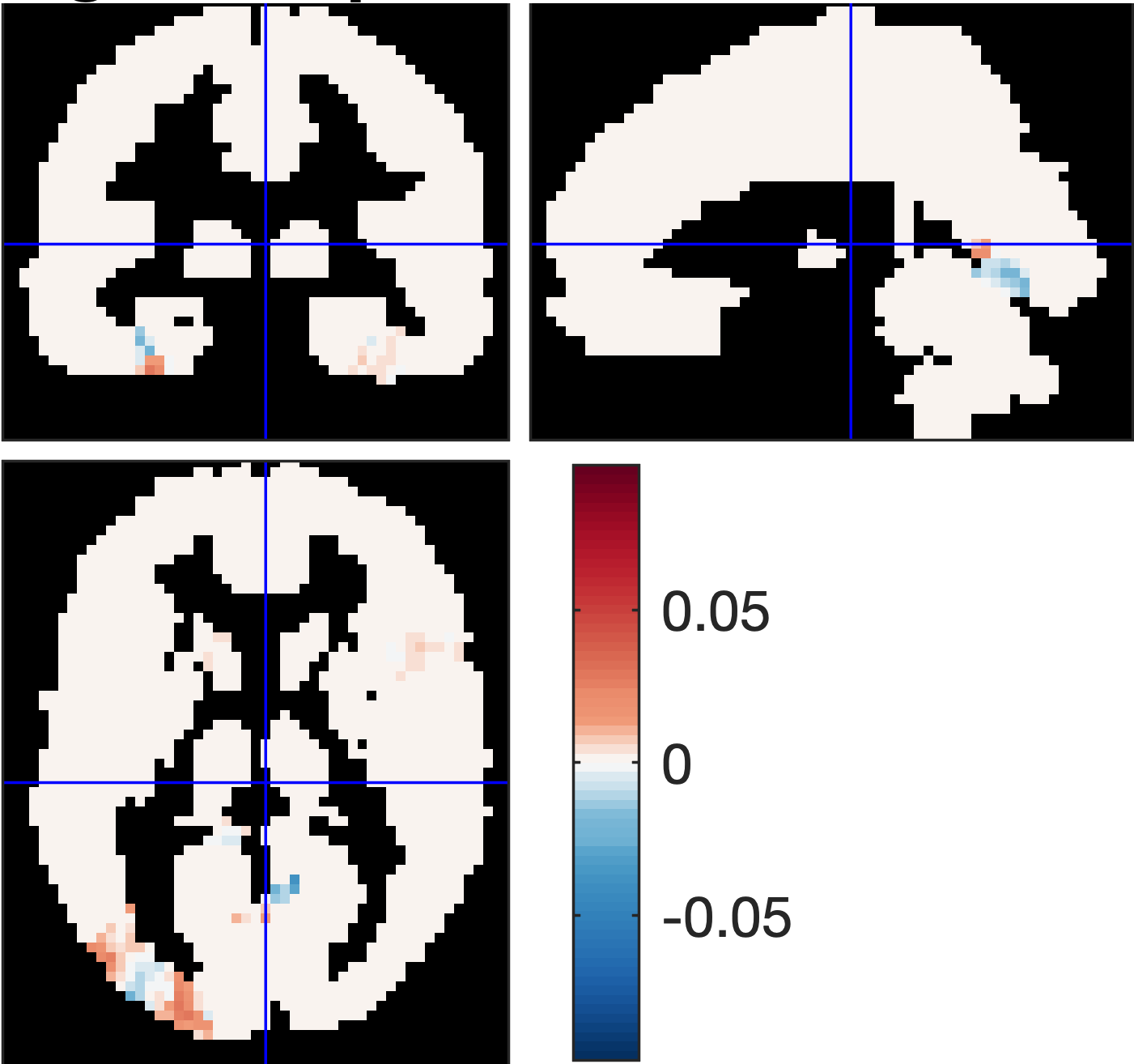}
\caption{SimpleMKL}
\end{subfigure}
~
\begin{subfigure}[b]{0.48\textwidth}
\centering
\includegraphics[width=\textwidth]{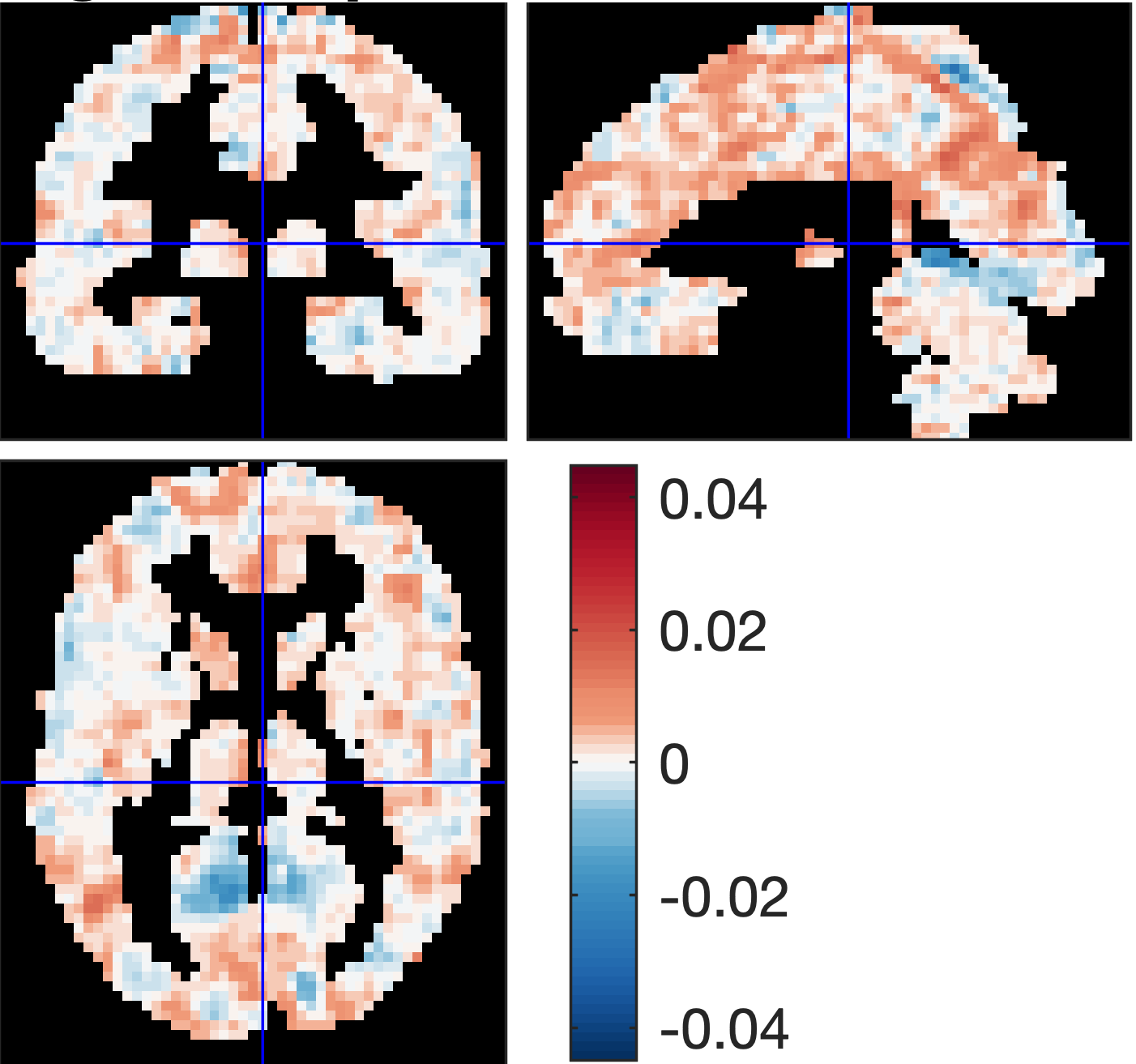}
\caption{SVM}
\end{subfigure}

\caption{Weight maps - Haxby data}
\label{fig:wmHaxby}

\end{figure}

\begin{figure}[H]
\centering
\begin{subfigure}[b]{0.48\textwidth}
\centering
\includegraphics[width=\textwidth]{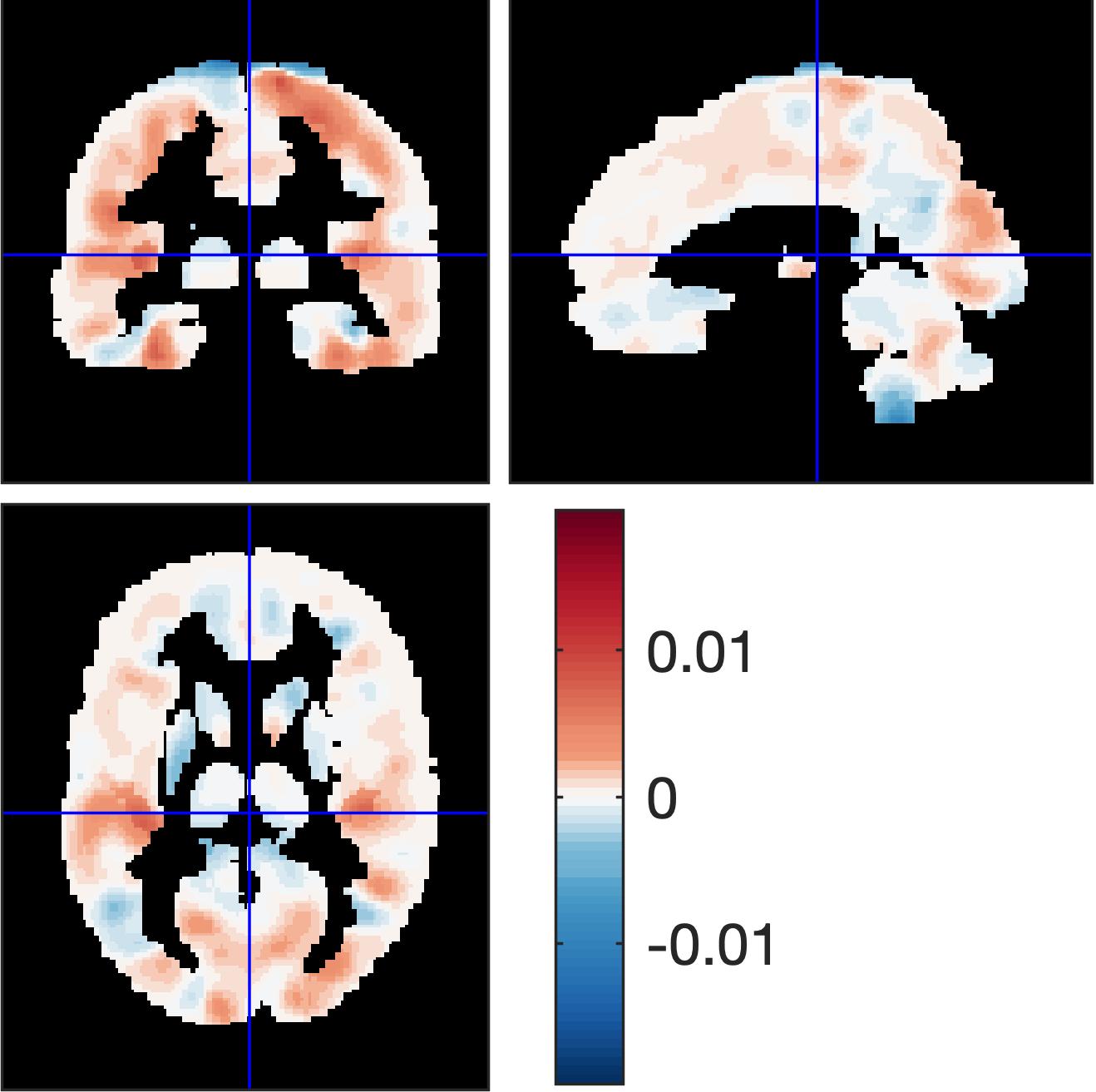}
\caption{ENMKL-SVM}
\end{subfigure}
~
\begin{subfigure}[b]{0.48\textwidth}
\centering
\includegraphics[width=\textwidth]{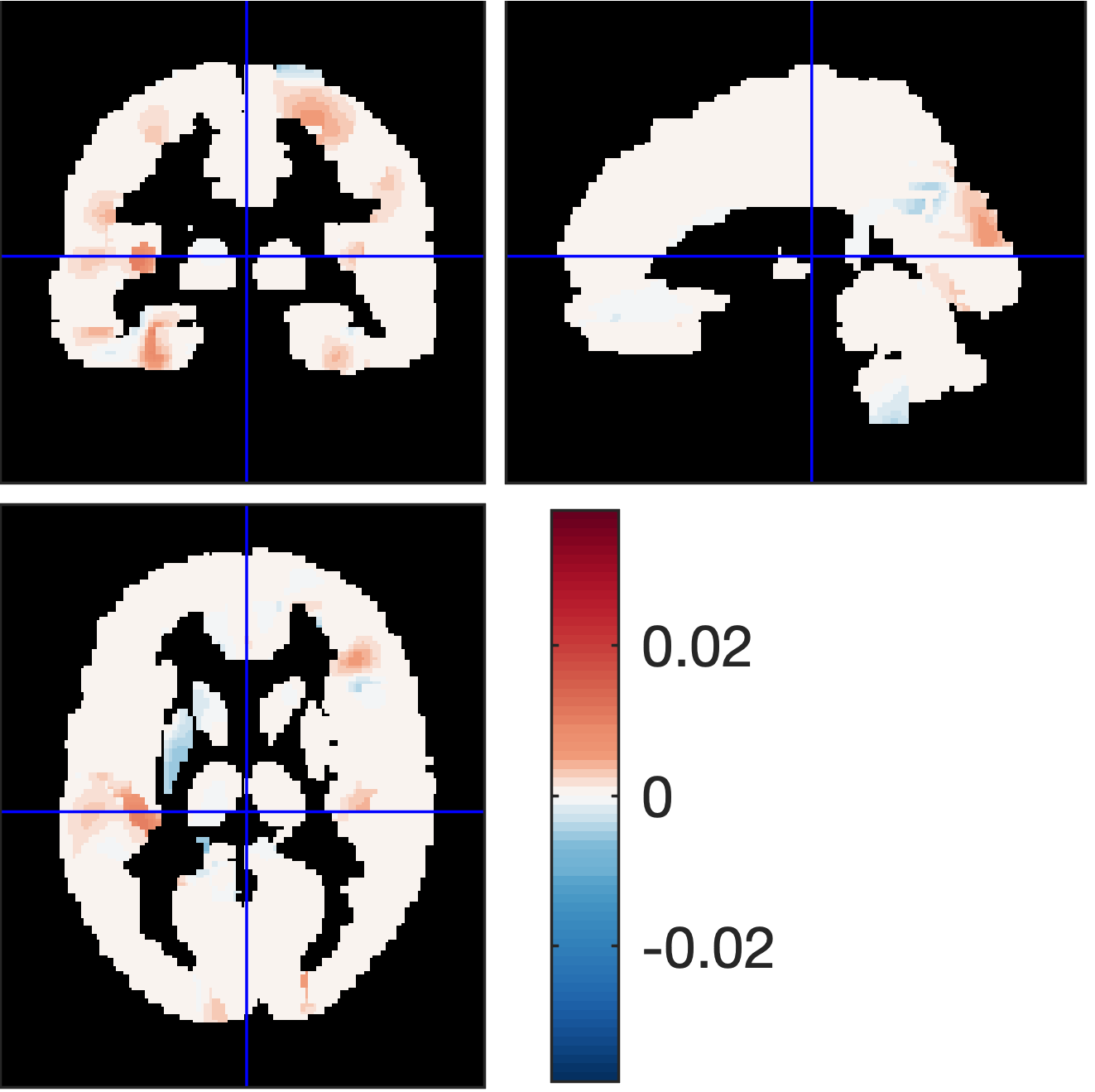}
\caption{SimpleMKL}
\end{subfigure}
~
\begin{subfigure}[b]{0.48\textwidth}
\centering
\includegraphics[width=\textwidth]{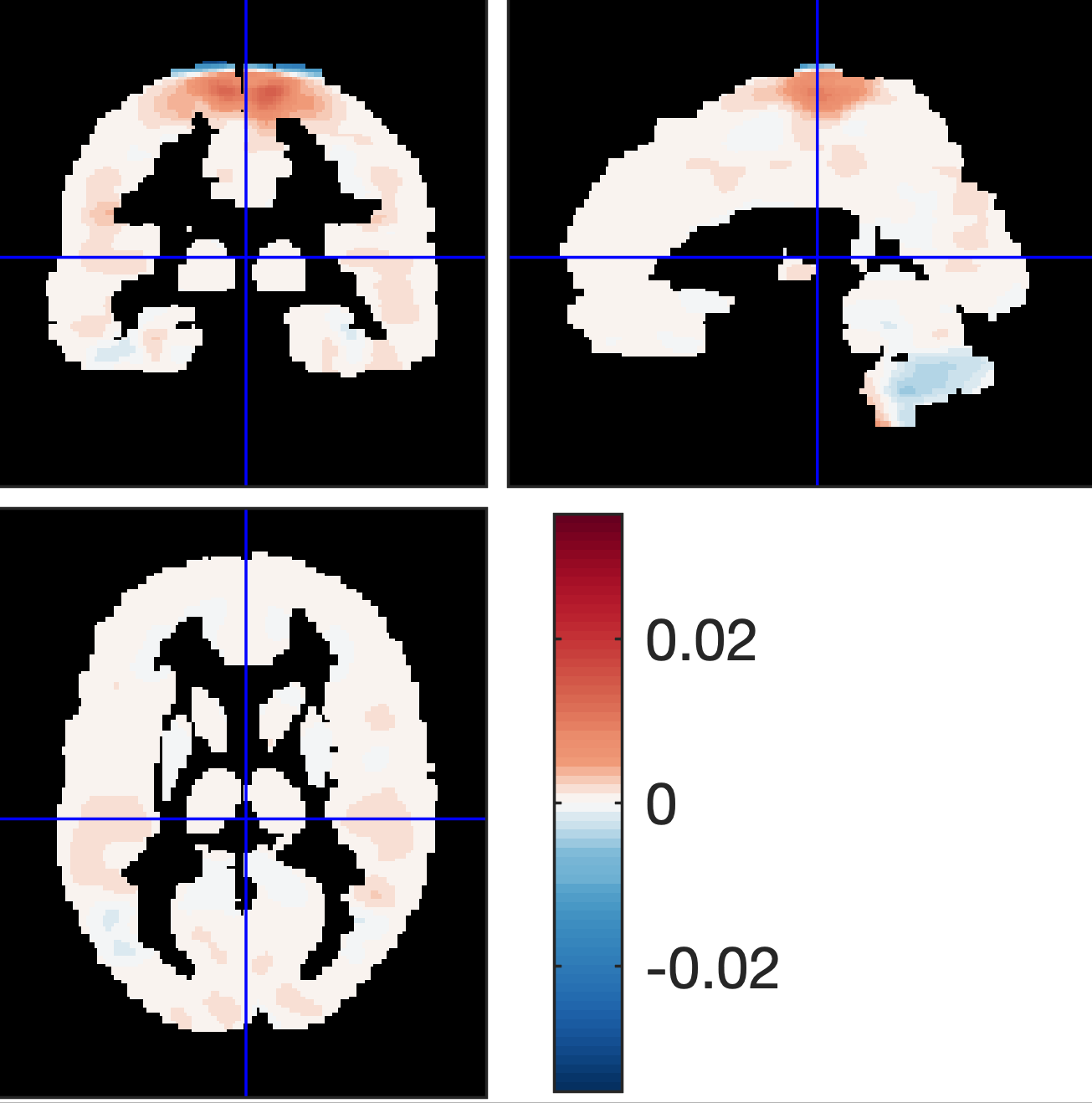}
\caption{KRR}
\end{subfigure}

\caption{Weight maps - IXI data}
\label{fig:wmIXI}

\end{figure}

The weight maps derived from the 3 datasets (OASIS, Haxby, and IXI) are depicted in Figures \ref{fig:wmOASIS}, \ref{fig:wmHaxby}, and \ref{fig:wmIXI} respectively, for the ENMKL-SVM, simple MKL and SVM/KRR solutions.


\clearpage

\clearpage

\clearpage

\section*{Acknowledgement}
J. Mourão-Miranda and K. Tsirlis were supported by the Wellcome Trust under Grant No. WT102845/Z/13/Z.
C. Phillips is a Research Director with the F.R.S.-FNRS (Belgian National Scientific Research Fund).


\bibliographystyle{plain} 
\bibliography{References}

@article{gonen2011multiple,
  title={Multiple Kernel Learning Algorithms},
  author={G{\"o}nen, Mehmet and Alpayd{\i}n, Ethem},
  journal={Journal of Machine Learning Research},
  volume={12},
  pages={2211--2268},
  year={2011}
}

@article{briscik2024supervised,
  title={Supervised multiple kernel learning approaches for multi‑omics data integration},
  author={Briscik, Mitja and Tazza, Gabriele and Vid{\'a}cs, L{\'a}szl{\'o} and Dillies, Marie‑Agn{\`e}s and D{\'e}jean, S{\'e}bastien},
  journal={BioData Mining},
  volume={17},
  number={1},
  pages={53},
  year={2024}
}

@article{fan2005working,
  author    = {Rong-En Fan and Pai-Hsuen Chen and Chih-Jen Lin},
  title     = {Working Set Selection Using Second Order Information for Training Support Vector Machines},
  journal   = {Journal of Machine Learning Research},
  volume    = {6},
  pages     = {1889--1918},
  year      = {2005},
  publisher = {JMLR.org}
}

@article{Aizerman1964,
  author  = {Aizerman, M.~A. and Braverman, E.~M. and Rozonoer, L.~I.},
  title   = {Theoretical Foundations of the Potential Function Method in Pattern Recognition Learning},
  journal = {Automation and Remote Control},
  volume  = {25},
  pages   = {821--837},
  year    = {1964},
}

@article{Zheng-Peng2011,
title = {{Elastic Multiple Kernel Learning}},
journal = {Acta Automatica Sinica},
volume = {37},
number = {6},
pages = {693-699},
year = {2011},
issn = {1874-1029},
doi = {https://doi.org/10.1016/S1874-1029(11)60208-5},
url = {https://www.sciencedirect.com/science/article/pii/S1874102911602085},
author = {Zheng-Peng Wu and Xue-Gong Zhang}
}

@inproceedings{Cortes2009,
title	= {{L2 Regularization for Learning Kernels}},
author	= {Corinna Cortes and Mehryar Mohri and Afshin Rostamizadeh},
year	= {2009},
URL	= {http://www.cs.nyu.edu/~mohri/postscript/l2reg-uai.pdf},
booktitle	= {Proceedings of the 25th Conference on Uncertainty in Artificial Intelligence  (UAI 2009)},
address	= {Montr\'eal, Canada}
}

@INBOOK{Buttou2007,
  author={Bottou, Léon and Chapelle, Olivier and DeCoste, Dennis and Weston, Jason},
  title={Large-scale kernel machines}, 
  year={2007},
  pages={1-27},
  publisher={MIT press}
}

@article{Kloft2008,
author = {Kloft, Marius and Brefeld, Ulf and Laskov, Pavel},
year = {2008},
month = {01},
pages = {},
title = {Non-sparse {Multiple Kernel Learning}},
journal = {NIPS 08 workshop: kernel learning automatic selection of optimal kernels}
}

@article{Bach2004,
author = {Bach, Francis R. and Lanckriet, Gert R. G. and Jordan, Michael I.},
doi = {10.1145/1015330.1015424},
isbn = {1581138385},
journal = {Proceedings, Twenty-First International Conference on Machine Learning, ICML 2004},
pages = {41--48},
title = {{Multiple kernel learning, conic duality, and the {SMO} algorithm}},
year = {2004}
}

@article{Citi2015,
author = {Citi, Luca},
doi = {10.1109/CEEC.2015.7332695},
isbn = {9781467394819},
journal = {2015 7th Computer Science and Electronic Engineering Conference, CEEC 2015 - Conference Proceedings},
pages = {29--34},
publisher = {IEEE},
title = {{Elastic-net constrained multiple kernel learning using a majorization-minimization approach}},
year = {2015}
}

@article{Kloft2011,
author  = {Marius Kloft and Ulf Brefeld and S{{\"o}}ren Sonnenburg and Alexander Zien},
  title   = {$lp$-{Norm Multiple Kernel Learning}},
  journal = {Journal of Machine Learning Research},
  year    = {2011},
  volume  = {12},
  number  = {26},
  pages   = {953--997},
  url     = {http://jmlr.org/papers/v12/kloft11a.html}
}

@article{Lanckriet2004,
author = {Lanckriet, Gert R.G. and Cristianini, Nello and Bartlett, Peter and {El Ghaoui}, Laurent and Jordan, Michael I.},
issn = {15337928},
journal = {Journal of Machine Learning Research},
pages = {27--72},
title = {{Learning the kernel matrix with semidefinite programming}},
volume = {5},
year = {2004}
}

@article{Micchelli2005,
author = {Micchelli, Charles A. and Pontil, Massimiliano},
issn = {15337928},
journal = {Journal of Machine Learning Research},
pages = {1099--1125},
title = {{Learning the kernel function via regularization}},
volume = {6},
year = {2005}
}

@article{Rakotomamonjy2008,
  author  = {Alain Rakotomamonjy and Francis R. Bach and St{{\'e}}phane Canu and Yves Grandvalet},
  title   = {{SimpleMKL}},
  journal = {Journal of Machine Learning Research},
  year    = {2008},
  volume  = {9},
  number  = {83},
  pages   = {2491--2521},
  url     = {http://jmlr.org/papers/v9/rakotomamonjy08a.html}
}

@article{Yang2011,
author = {Yang, Haiqin and Xu, Zenglin and Ye, Jieping and King, Irwin and Lyu, Michael R.},
doi = {10.1109/TNN.2010.2103571},
issn = {10459227},
journal = {IEEE Transactions on Neural Networks},
number = {3},
pages = {433--446},
pmid = {21257374},
publisher = {IEEE},
title = {{Efficient sparse generalized multiple kernel learning}},
volume = {22},
year = {2011}
}

@article{tzourio2002automated,
  title={Automated anatomical labeling of activations in {SPM} using a macroscopic anatomical parcellation of the {MNI} {MRI} single-subject brain},
  author={Tzourio-Mazoyer, Nathalie and Landeau, Brigitte and Papathanassiou, Dimitri and Crivello, Fabrice and Etard, Octave and Delcroix, Nicolas and Mazoyer, Bernard and Joliot, Marc},
  journal={Neuroimage},
  volume={15},
  number={1},
  pages={273--289},
  year={2002},
  publisher={Elsevier}
}

@book{shawe2004kernel,
  title={Kernel methods for pattern analysis},
  author={Shawe-Taylor, John and Cristianini, Nello and others},
  year={2004},
  publisher={Cambridge university press}
}

@article{haxby2001distributed,
  title={Distributed and overlapping representations of faces and objects in ventral temporal cortex},
  author={Haxby, James V. and Gobbini, M. Ida and Furey, Maura L. and Ishai, Alumit and Schouten, Jennifer L. and Pietrini, Pietro},
  journal={Science},
  volume={293},
  number={5539},
  pages={2425--2430},
  year={2001},
  publisher={American Association for the Advancement of Science}
}

@article{schrouff2013pronto,
  title={{PRoNTo}: {P}attern {R}ecognition for {N}euroimaging {T}oolbox},
  author={Schrouff, Jessica and Rosa, Maria J and Rondina, Jane M and Marquand, Andre F and Chu, Carlton and Ashburner, John and Phillips, Christophe and Richiardi, Jonas and Mourao-Miranda, Janaina},
  journal={Neuroinformatics},
  volume={11},
  number={3},
  pages={319--337},
  year={2013},
  publisher={Springer}
}

@inproceedings{argyriou2005learning,
  title={Learning convex combinations of continuously parameterized basic kernels},
  author={Argyriou, Andreas and Micchelli, Charles A and Pontil, Massimiliano},
  booktitle={International Conference on Computational Learning Theory},
  pages={338--352},
  year={2005},
  organization={Springer}
}

\end{document}